\documentclass[journal,onecolumn,10pt]{IEEEtran}

\pdfoutput=1
\usepackage{amsmath}
\usepackage{amssymb}
\usepackage{amsthm}
\usepackage{amsfonts}
\usepackage{epsfig}
\usepackage{algorithm2e}
\usepackage{xcolor}
\usepackage{multirow}
\usepackage{textcomp}

\renewcommand{\theequation}{\thesection.\arabic{equation}}

\theoremstyle{definition}
\newtheorem{theorem}{Theorem}
\newtheorem{proposition}{Proposition}
\newtheorem{lemma}{Lemma}

\newcommand{\BlackBox}{\rule{1.5ex}{1.5ex}}  
\newenvironment{proofi}{\par\noindent{\bf Proof: }}{\hfill\BlackBox\\}

\def\reals{\ensuremath{\mathbb{R}}}
\def\naturals{\ensuremath{\mathbb{N}}}

\newcommand{\mat}[1]{\ensuremath{\mathbf{{#1}}}}
\renewcommand{\vec}[1]{\ensuremath{\mathbf{\MakeLowercase{#1}}}}
\newcommand{\subjto}{\ensuremath{\quad\mathrm{s.t.}\quad}}
\newcommand{\norm}[1]{\ensuremath{\left\|#1\right\|}}
\newcommand{\cost}[1]{\ensuremath{\ell_{#1}}}

\newcommand{\setdef}[1]{\ensuremath{\left\{#1\right\}}}

\def\transp{^{\intercal}}
\newcommand{\opt}[1]{{#1}^*} \newcommand{\refeq}[1]{(\ref{#1})}

\newcommand{\iter}[1]{^{(#1)}}
\def\sparsa{\textsc{s}pa\textsc{rsa}}
\def\defeq{:=}

\newcommand{\sblock}[1]{_{[#1]}}

\def\dictm{\mat{D}}
\def\dictv{\mat{d}}

\def\datam{\mat{X}}
\def\datav{\mat{x}}

\def\coefm{\mat{A}}
\def\coefv{\mat{a}}
\def\coef{a}

\def\auxm{\mat{Z}}
\def\auxv{\mat{z}}

\newcommand{\inv}[1]{\left(#1\right)^{-1}}
\def\pinv{\mat{H}}
\def\oinv{\mat{Q}}

\def\ndims{m}
\def\natoms{p}
\def\nsamples{n}
\def\group{G}
\def\groupset{\mathcal{G}}
\def\ngroups{q}
\def\reg{\psi}
\def\sgn{\mathrm{sgn}}
\def\groupsize{g}

\def\mutualco{{\mu}}
\def\blockco{{\mu_{B}}}
\def\subco{{\nu}}
\def\crossco{{\chi}}
\def\regG{\reg_{\groupset}}
\newcommand\rhoc[1]{\rho_{[#1]}}
\newcommand{\cpl}[1]{\overline{#1}} 

\newcommand{\bv}{{\bf v}}

\newcommand{\bbi}{{\bf I}}

\newcommand{\bbw}{{\bf W}}

\newcommand{\blc}{\left\{}
\newcommand{\brc}{\right\}}

\begin{document}
\title{ {\it C-HiLasso:} A Collaborative Hierarchical Sparse Modeling Framework}

\author{Pablo Sprechmann,$^1$$^\dagger$ Ignacio Ram\'{i}rez,$^1$$^\dagger$ Guillermo
  Sapiro$^1$ and Yonina C. Eldar$^2$ \\ $^1$University of Minnesota and
  $^2$Technion\thanks{$^\dagger$P. S. and I. R. contributed equally to this
    work.}}

\maketitle
\begin{abstract}
  Sparse modeling is a powerful framework for data analysis and
  processing. Traditionally, encoding in this framework is performed by
  solving an $\ell_1$-regularized linear regression problem, commonly
  referred to as \emph{Lasso} or \emph{Basis Pursuit}. In this work we
  combine the sparsity-inducing property of the Lasso at the individual
  feature level, with the block-sparsity property of the \emph{Group Lasso},
  where sparse groups of features are jointly encoded, obtaining a sparsity
  pattern hierarchically structured. This results in the \emph{Hierarchical
    Lasso (HiLasso)}, which shows important practical advantages.  We then
  extend this approach to the collaborative case, where a set of
  simultaneously coded signals share the same sparsity pattern at the higher
  (group) level, but not necessarily at the lower (inside the group) level,
  obtaining the collaborative HiLasso model \emph{(C-HiLasso)}. Such signals
  then share the same active groups, or classes, but not necessarily the
  same active set. This model is very well suited for applications such as
  source identification and separation. An efficient optimization procedure,
  which guarantees convergence to the global optimum, is developed for these
  new models. The underlying presentation of the framework and optimization
  approach is complemented by experimental examples and theoretical results
  regarding recovery guarantees.
\end{abstract}

\section{Introduction and Motivation} 

Sparse signal modeling has been shown to lead to numerous state-of-the-art results in
signal processing, in addition to being very attractive at the theoretical level. The standard model assumes that a signal can be
efficiently represented by a sparse linear combination of atoms from a given
or learned dictionary. The selected atoms form what is usually referred to
as the {\it active set}, whose cardinality is significantly smaller than the
size of the dictionary and the dimension of the signal.

In recent years, it has been shown that adding structural
constraints to this active set has value both at the level of
representation robustness and at the level of signal
interpretation (in particular when the active set indicates some
physical properties of the signal); see  \cite{yuan06,bach09,EM09}
and references therein. This leads to {\it group} or {\it
structured} sparse coding, where instead of considering the atoms
as singletons, the atoms are grouped, and a few groups are active
at a time. An alternative way to add structure (and robustness) to
the problem is to consider the simultaneous encoding of multiple
signals, requesting that they all share the same active set. This
is a natural collaborative filtering approach to sparse coding;
see, for example, \cite{tropp06a,tropp06b,CREK05,CH06,ME08a,ER10}.

In this work we extend these approaches in a number of directions. First, we
present a hierarchical sparse model, where not only a few (sparse) groups of
atoms are active at a time, but also each group enjoys internal
sparsity.\footnote{While we consider only 2 levels of sparsity, the
  proposed framework is easily extended to multiple levels.} At the
conceptual level, this means that the signal is represented by a few groups
(classes), and inside each group only a few members are active at a time. A
simple example of this is a piece of music (numerous applications in
genomics and image processing exist as well), where only a few instruments
are active at a time (each instrument is a group), and the sound produced by
each instrument at each instant is efficiently represented by a few atoms of
the sub-dictionary/group corresponding to it. Thereby, this proposed
hierarchical sparse coding framework permits to efficiently perform source
identification and separation, where the individual sources (classes/groups)
that generated the signal are identified at the same time as their
representation is reconstructed (via the sparse code inside the group). An
efficient optimization procedure, guaranteed to converge to the global
optimum, is proposed to solve the hierarchical sparse coding problems that
arise in our framework. Theoretical recovery bounds are derived, which guarantee that the output of the optimization algorithm is the true underlying signal.

Next, we go one step beyond. Continuing with the above example, if we know
that the same few instruments will be playing simultaneously during
different passages of the piece, then we can assume that the active groups
at each instant, within the same passage, will be the same.  We can exploit
this information by applying the new hierarchical sparse coding approach in
a collaborative way, enforcing that the same groups will be active at all
instants within a passage (since they are of the same instruments and then
efficiently representable by the same sub-dictionaries), while allowing each
group for each music instant to have its own unique internal sparsity
pattern (depending on how the sound of each instrument is represented at
each instant). We propose a collaborative hierarchical sparse coding
framework following this approach, {\it(C-HiLasso)}, %
along with an efficient
optimization procedure. We then comment on results regarding the correct
recovery of the underlying active groups.

The proposed optimization techniques for both \emph{HiLasso} and \emph{C-HiLasso} is based on the Proximal Method~\cite{nesterov07}, more
  specifically, on its particular implementation for sparse problems,
  \emph{Sparse Reconstruction by Separable Approximation} (\sparsa)
  \cite{SpaRSA}. This is an iterative method which solves a subproblem at
  each iteration which, in our case, has a closed form and can be solved in
  linear time. Furthermore, this closed form solution combines a vector
  thresholding and a scalar thresholding, naturally yielding to the desired
  hierarchical sparsity patterns.

The rest of the paper is organized as follows: Section~\ref{sec:intro}
provides an introduction to traditional sparse modeling and presents our
proposed HiLasso and C-HiLasso models. We discuss their relationship with
the recent works of \cite{bach09,friedman10a,peng,KimICML,Jenatton2010,
  Starck04imagedecomposition}. In Section~\ref{sec:opt} we describe the
optimization techniques applied to solve the resulting sparse coding
problems and we discuss its relationship with other optimization methods
recently proposed in the literature
\cite{Jenatton2010,JenattonBach}. Theoretical recovery guarantees for
HiLasso in the noiseless setting are developed in Section~\ref{sec:theory},
demonstrating improved performance when compared with Lasso and Group
Lasso. We also comment on existing results regarding correct recovery of
group-sparse patterns in the collaborative case. Experimental results and
simulations are given in Section~\ref{sec:results}, and finally concluding
remarks are presented in Section~\ref{sec:discussion}.

\section{Collaborative Hierarchical Sparse Coding}
\label{sec:intro}

\subsection{Background: Lasso and Group Lasso}

Assume we have a set of data samples $\datav_j \in
\reals^{\ndims}, j=1,\ldots,\nsamples$, and a dictionary of
$\natoms$ atoms in $\reals^{\ndims}$, assembled as a matrix $\dictm \in
\reals^{\ndims{\times}\natoms}$, $\dictm=[\dictv_1 \dictv_2 \ldots
\dictv_\natoms]$. Each sample $\datav_j$ can be written as
$\datav_j=\dictm\coefv_j+\epsilon,\,\coefv_j \in
\reals^{\natoms},\,\epsilon \in \reals^{\ndims}$, that is, as a
linear combination of the atoms in the dictionary $\dictm$ plus
some perturbation $\epsilon$, satisfying $\norm{\epsilon}_2\ll \norm{\datav_j}_2$. The basic underlying assumption in
sparse modeling is that, for all or most $j$, the ``optimal''
 $\coefv_j$ has only a few nonzero elements.
Formally, if we define the $\cost{0}$ cost  as the pseudo-norm
counting the number of nonzero elements of $\coefv_j$,
$\norm{\coefv_j}_0\defeq|\{k:\coef_{kj} \neq 0\}|$, then we expect
that $\norm{\coefv_j}_0 \ll \natoms$ and $\norm{\coefv_j}_0 \ll \ndims $  for all or most $j$.

Seeking the sparsest representation $\coefv$ is known to be
\textsc{NP}-hard. To determine $\coefv_j$ in practice, a multitude of
efficient algorithms have been proposed, which achieve high correct recovery
rates. The $\ell_1$-minimization method is the most extensively studied
recovery technique. In this approach, the non-convex $\ell_0$ norm is
replaced by the convex $\ell_1$ norm, leading to
\begin{equation}
\label{eq:bp} \min_{\coefv\in\reals^{\natoms}} \norm{\coefv}_1 \subjto
\norm{\datav_j-\dictm\coefv}_2^2 \leq \epsilon.
\end{equation}
 The use of general
purpose or specialized convex optimization techniques allows for
efficient reconstruction using this strategy. The above
approximation is known as the Lasso \cite{tibshirani96} or Basis Pursuit \cite{CDS99,D06}. 
A popular variant is to use the unconstrained version
\begin{equation}
\min_{\coefv\in\reals^{\natoms}} \frac{1}{2}\norm{\datav_j-\dictm\coefv}_2^2 + \lambda\norm{\coefv}_1,
\label{eq:lasso}
\end{equation}
where $\lambda$ is an appropriate parameter value, usually found by
cross-validation, or based on statistical principles~\cite{giryes10}.

The fact that the $\norm{\cdot}_1$ regularizer induces sparsity in the
solution $\coefv_j$ is desirable not only from a regularization point of
view, but also from a model selection perspective, where one wants to
identify the relevant factors (atoms) that conform each sample
$\datav_j$. In many situations, however, the goal is to represent the
relevant factors not as singletons but as groups of atoms. For a dictionary
of $\natoms$ atoms, we define groups of atoms through their indices, $G
\subseteq \{1,\ldots,\natoms\}$. Given a group $G$ of indexes, we denote the
sub-dictionary of the columns indexed by them as $\dictm\sblock{G}$, and the
corresponding set of reconstruction coefficients as
$\coefv\sblock{G}$. Define $\groupset = \{G_1,\ldots,G_\ngroups \}$ to be a
partition of $\{1,\ldots,\natoms\}$.\footnote{While in this paper we
  concentrate and develop the important non-overlapping case, it will be
  clear that the concepts of collaborative hierarchical sparse modeling
  introduced here apply to the case of overlapping groups as well.} In order
to perform model selection at the group level (relative to the partition
$\groupset$), the Group Lasso problem was introduced in \cite{yuan06},
\begin{equation}
\min_{\coefv\in\reals^{\natoms}} \frac{1}{2}\norm{\datav_j-\dictm\coefv}_2^2 + \lambda\regG(\coefv),
\label{eq:group-lasso}
\end{equation}
where $\regG$ is the Group Lasso regularizer defined in terms of $\groupset$
as $\regG(\coefv) \defeq \sum_{\group \in
  \groupset}{\norm{\coefv\sblock{G}}_2}$. The function $\regG$ can be seen
as a generalization of the $\cost{1}$ regularizer, as the latter arises from
the special case $\groupset=\setdef{\{1\},\{2\},\ldots,\{\natoms\}}$ (the
groups are singletons), and as such, its effect on the groups of $\coefv$ is
also a natural generalization of the one obtained with the Lasso: it ``turns
on/off'' atoms in groups.


We can always consider the ``noiseless'' sparse coding problem
$\min_{\coefv\in\reals^{\natoms}}\setdef{ \reg(\coefv): \datav_j=\dictm\coefv}$, for a
generic regularizer $\reg(\cdot)$, as the limit of the Lagrangian sparse
coding problem $\min_{\coefv\in\reals^{\natoms}}\setdef{
  \frac{1}{2}\norm{\datav_j-\dictm\coefv}_2^2 + \lambda\reg(\coefv)}$ when
$\lambda \rightarrow 0$. In the remainder of this section, as well as in
Section~\ref{sec:opt}, we only present the corresponding Lagrangian
formulations.


\subsection{The Hierarchical Lasso}
\label{sec:background:Hilasso}

\begin{figure*}
\begin{center}
\includegraphics[width=0.45\textwidth]{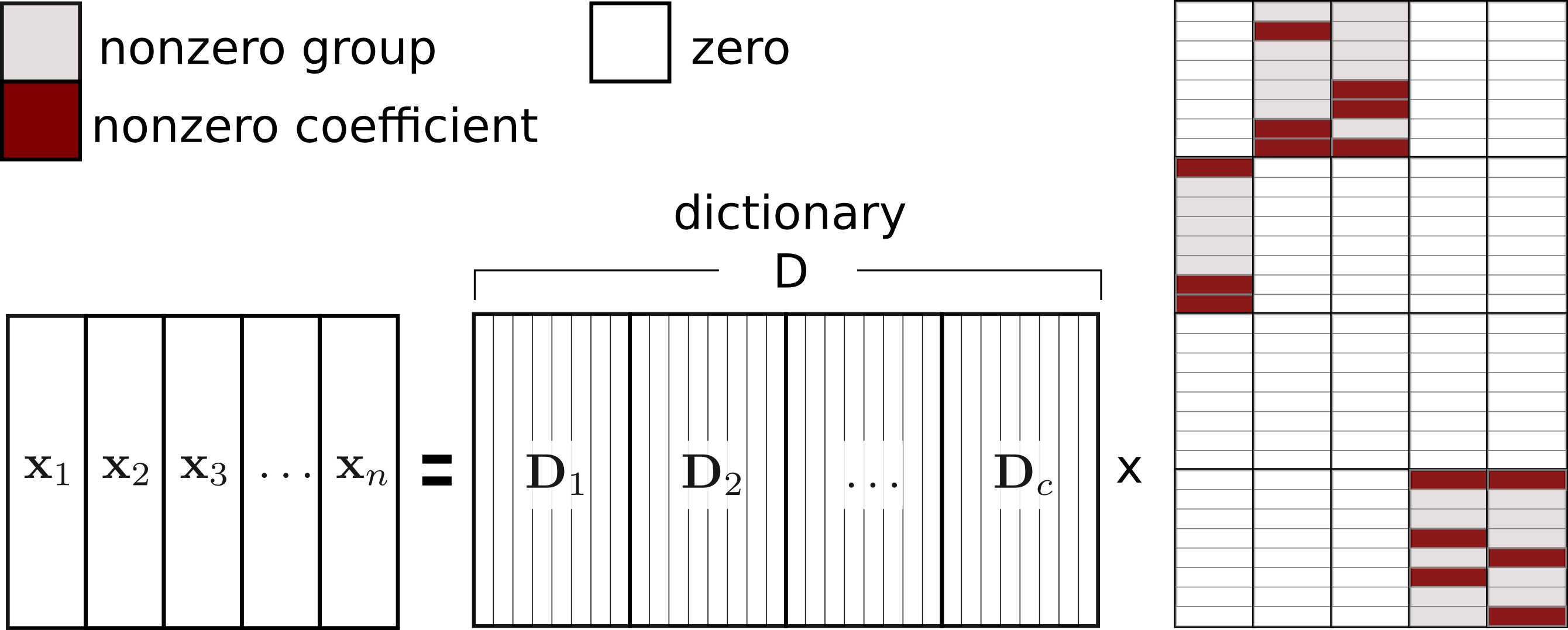}\hspace{8ex}%
\includegraphics[width=0.45\textwidth]{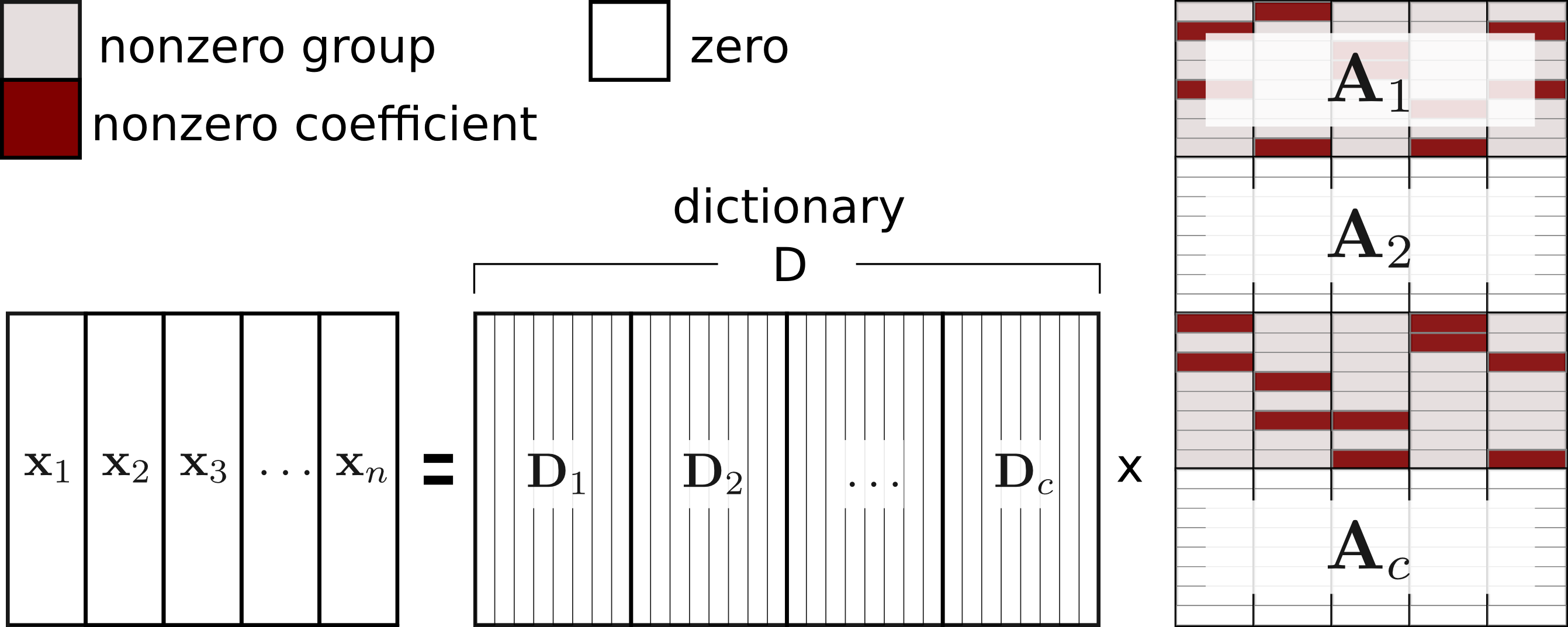}
\vspace{-1ex}\caption{%
  \label{fig:models}%
  \footnotesize%
  Sparsity patterns induced by HiLasso (left) and
  C-HiLasso (right) model selection programs. Notice that the
  C-HiLasso imposes the same group-sparsity pattern in all the samples (same class),
  whereas the in-group sparsity patterns can vary between
  samples (samples themselves are different). }
\end{center}
\end{figure*}

The Group Lasso trades sparsity at the single-coefficient level with
sparsity at a group level, while, inside each group, the solution is
generally dense. Let us consider for example that each group is a
sub-dictionary trained to efficiently represent, via sparse modeling, an
instrument, a type of image, or a given class of signals in general. The
entire dictionary $\dictm$ is then appropriate to represent all classes of
the signal as well as mixtures of them, and Group Lasso will properly
represent (dense) mixtures with one group or sub-dictionary per class. At
the same time, since each class is properly represented in a sparse mode via
its corresponding group or sub-dictionary, we expect sparsity inside its
groups as well (which is not achieved by Group Lasso, whose solutions are
dense inside each group). This will become even more critical in the
collaborative case, where signals will share groups because they are of the
same class, but will not necessarily share the full active sets, since they
are not the same signal. To achieve the desired in-group sparsity, we simply
re-introduce the $\cost{1}$ regularizer together with the group regularizer,
leading to the proposed {\it Hierarchical Lasso (HiLasso)}
model,\footnote{We can similarly define a hierarchical sparsity model with
  $\ell_0$ instead of $\ell_1$.}
\begin{equation}
\min_{\coefv\in\reals^{\natoms}} \frac{1}{2}\norm{\datav_j-\dictm\coefv}_2^2 +
\lambda_2\regG(\coefv) + \lambda_1\norm{\coefv}_1.
\label{eq:hilasso}
\end{equation}
The hierarchical sparsity pattern produced by the solutions of
\refeq{eq:hilasso} is depicted in Figure~\ref{fig:models}(left). For
simplicity of the description, we assume that all the groups have the same
number of elements. The extension to the general case is obtained by
multiplying each group norm by the square root of the corresponding group
size. This model then achieves the desired effect of promoting sparsity at
the group/class level while at the same time leading to overall sparse
feature selection. As mentioned above, additional levels of hierarchy can be
considered as well, e.g., with groups inside the blocks. This is relevant
for example in audio analysis.

\begin{figure*}
\begin{center}\includegraphics[width=0.8\textwidth]{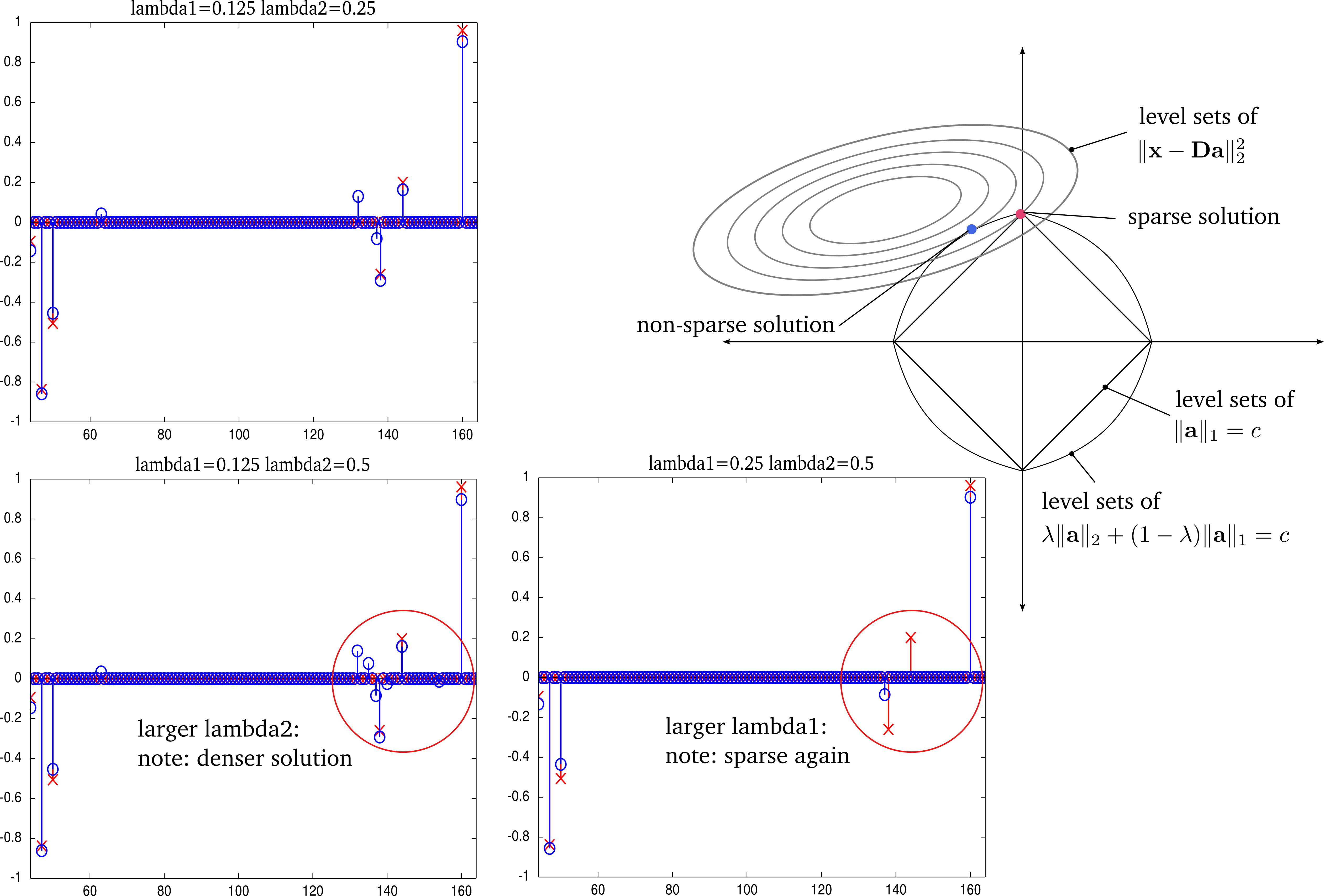}\end{center}
\vspace{-3ex}\caption{\label{fig:lambda-effect}Effect of different
  combinations of $\lambda_1$ and $\lambda_2$ on the solutions of the
  HiLasso coding problem. Three cases are given in which we want to recover
  a sparse signal (red crosses) $\coefv_0$ by means of the solution $\coefv$
  of the HiLasso problem (blue dots). In this example we have two active
  groups out of ten possible (the sub dictionaries associated to each group
  have 30 atoms) and $\coefv_0=8$ (four non-zero coefficient per active
  group). The estimate that is closest to $\coefv_0$ in $\cost{1}$ norm is
  shown in the top left. As the ratio $\lambda_2/\lambda_1$ increases
  (bottom left), the level sets of the regularizer $\regG(\cdot)$ become
  rounder, thus encouraging denser solutions. This is depicted in the
  rightmost figure for a simple case of $\ngroups=1$ groups. Increasing
  $\lambda_1$ again (bottom right) increases sparsity, although here the
  final effect is too strong and some non-zero coefficients are not
  detected. }
\end{figure*}

As with models such as Lasso and Group Lasso, the optimal parameters
$\lambda_1$ and $\lambda_2$ are application and data dependent. In some
specific cases, closed form solutions exist for such parameters. For
example, for signal restoration in the presence of noise using Lasso
($\lambda_2=0$), the \textsc{gsure} method provides a simple way to compute
the optimal $\lambda_1$ \cite{giryes10}. As extending such methods to
HiLasso (or the C-HiLasso model presented next) is beyond the scope of this
work, we rely on cross-validation for the choice of such parameters. The
selection of $\lambda_1$ and $\lambda_2$ has an important influence on the
sparsity of the obtained solution. Intuitively, as $\lambda_2/\lambda_1$
increases, the group constraint becomes dominant and the solution tends to
be more sparse at a group level but less sparse within groups (see
Figure~\ref{fig:lambda-effect}). This relation allows in practice to
intuitively select a set of parameters that performs well. We also noticed
empirically that the selection of the parameters is quite robust, since
small variations in their numerical value don't change considerably the
obtained results.

Some recent modeling frameworks for sparse coding do not rely on the
selection of such model parameters, e.g., following the Minimum Description
Length criterion in \cite{nachoMDL}, or non-parametric Bayesian techniques
in \cite{zhouNIPS}. Applying such techniques to the here proposed
  models is subject of future research.

\subsection{Collaborative Hierarchical Lasso}
\label{sec:background:collaborative-lasso}

In numerous applications, one expects that certain collections of
samples $\datav_j$ share the same active components from the
dictionary, that is, that the indices of the nonzero coefficients
in $\coefv_j$ are the same for all the samples in the collection.
Imposing such dependency in the $\cost{1}$ regularized regression
problem gives rise to the so called collaborative (also called
``multitask'' or ``simultaneous'') sparse coding problem
\cite{tropp06a,ME08a,ER10,wright04}. Considering the coefficients matrix 
$\coefm=[\coefv_1,\ldots,\coefv_\nsamples]\in \reals^{\natoms{\times}
  \nsamples}$ associated with the reconstruction of the samples
$\datam=[\datav_1,\ldots,\datav_\nsamples]\in \reals^{\ndims{\times}
  \nsamples}$, this model is given by
\begin{equation}
\min_{\coefm\in\reals^{\natoms{\times}\nsamples}} \frac{1}{2}\norm{\datam - \dictm\coefm}_F^2 +
 \lambda \sum_{k=1}^{\natoms}{\norm{\coefv^k}_2},
\label{eq:collaborative-lasso}
\end{equation}
where $\coefv^k\in \reals^{\nsamples}$ is the $k$-th row of $\coefm$, that
is, the vector of the $\nsamples$ different values that the coefficient
associated to the $k$-th atom takes for each sample $j=1,\ldots,\nsamples$.
If we now extend this idea to the Group Lasso, we obtain a {\it
  collaborative Group Lasso} (\emph{C-GLasso}) formulation,
\begin{equation}
\min_{\coefm\in\reals^{\natoms{\times}\nsamples}} \frac{1}{2}\norm{\datam- \dictm\coefm}_F^2 +
 \lambda\regG(\coefm),
\label{eq:collaborative-group-lasso}
\end{equation}
where $\regG(\coefm) = \sum_{\group \in
  \groupset}{\norm{\coefm^\group}_F}$, and $\coefm^\group$ is the sub-matrix
formed by all the rows belonging to group $\group$.  This regularizer is the
natural collaborative extension of the regularizer in \refeq{eq:group-lasso}.

In this paper, we take an additional step and treat this together with the
hierarchical extension presented in the previous section. The combined model
that we propose, \emph{C-HiLasso}, is given by
\begin{equation}
\min_{\coefm\in\reals^{\natoms{\times}\nsamples}} \frac12 \norm{\datam - \dictm\coefm}_F^2
+   \lambda_2 \regG(\coefm) + \lambda_1 \sum_{j=1}^{\nsamples}  \norm{\coefv_j}_1.
\label{eq:collaborative-hilasso}
\end{equation}
The sparsity pattern obtained using \refeq{eq:collaborative-hilasso} is
shown in Figure~\ref{fig:models}(right). The C-GLasso is a particular case
of our model when $\lambda_1=0$. On the other hand, one can obtain
independent Lasso solutions for each $\mat{x}_i$ by setting $\lambda_2=0$.
We see that \refeq{eq:collaborative-hilasso} encourages all the signals to
share the same groups (classes), while the active set inside each group is
signal dependent. We thereby obtain a collaborative hierarchical sparse
model, with collaboration at the class level (all signals collaborate to
identify the classes), and freedom at the individual levels inside the class
to adapt to each particular signal. This new model is particularly well
suited, for example, when the data vectors have missing components. In this
case combining the information from all the samples is very important in
order to obtain a correct representation and model (group) selection. This
can be done by slightly changing the data term in
\refeq{eq:collaborative-hilasso}.  For each data vector $\datav_j$ one
computes the reconstruction error using only the observed elements. Note
that the missing components do not affect the other terms of the
equation. Examples will be shown in Section~\ref{sec:results}.


\subsection{Relationship to Recent Literature}
\label{sec:background:literature}

A number of recent works have addressed hierarchy, grouping and
collaboration within the sparse modeling community. We now discuss the ones
most closely related to the proposed HiLasso and C-HiLasso models.

In \cite{bach09}, the authors propose a general framework in which one can
define a regularization term to encourage a variety of sparsity patterns,
and provide theoretical results (different to the ones developed here) for
the single-signal case. The HiLasso model presented here, in the single
signal scenario, can be seen as a particular case of that model (where the
groups in \cite{bach09} should be blocks and singletons), although the
particularly and important case of hierarchical structure introduced here is
not mentioned in that paper. In \cite{friedman10a} the authors
simultaneously (see \cite{CISS2010}) proposed a model that coincides with
ours again in the single-signal scenario. None of these approaches develop
the collaborative framework introduced here, nor the theoretical guarantees.
The recovery of mixed signals with $\ell_0$ optimization was addressed in
\cite{Starck04imagedecomposition}. This model does not include block
sparsity (no hierarchy), collaboration, or the theoretical results we
obtain here.

The special case of C-HiLasso when $\lambda_1=0$, C-GLasso, is investigated
in \cite{boufounos10}, where a theoretical analysis of the signal recovery
properties of the model is developed. Collaborative coding with structured
sparsity has also been used recently in the context of gene expression
analysis \cite{peng,KimICML}. In \cite{peng}, the authors propose a model,
that can be interpreted as a particular case of the collaborative approach
presented here, in which a set of signals is simultaneously coded using a
small (sparse) number of atoms of the dictionary. They modify the classical
collaborative sparse coding regularization so that each signal can use any
subset of the detected atoms.  This is equivalent to our model when the
groups have only one element and therefore there is no hierarchy in the
coding. %
A collaborative model is presented in \cite{KimICML}, where signals sharing
the same active atoms are grouped together in a hierarchical way by means of
a tree structure.  The regularization term proposed is analogous to the one
proposed in our work, but it is used to group signals rather than atoms
(features), having once again no hierarchical coding.

Tree-based sparse coding has also been used recently to learn
dictionaries~\cite{Jenatton2010,JenattonBach}. Under this model, if a
particular learned atom is not used in the decomposition of a signal, then
none of its descendants (in terms of the given tree structure) can be
used. Although not explicitly considered in these works, the HiLasso model
is an important particular case, among the wide spectrum of hierarchical
sparse models considered in this line of work, where the hierarchy has two
levels and no single atoms are in the upper level.

To conclude, while particular instances of the proposed C-HiLasso have been
recently reported in the literature, none of them are as
comprehensive. C-HiLasso includes both collaboration, at a block/group
level, and hierarchical coding. Such collaborative hierarchical structure is
novel and fundamental to address new important problems such as
collaborative source identification and separation. The new theoretical
results presented here extend the block sparsity results of
\cite{EM09,EKB09}, complementing the modeling and algorithmic work.


\section{Optimization}
\label{sec:opt}

\subsection{Single-Signal Problem: HiLasso}
\label{sec:opt.single}

In the last decade, optimization of problems of the form of \eqref{eq:lasso}
and \eqref{eq:group-lasso} have been deeply studied, and there exist very
efficient algorithms for solving them. Recently, Wright et. al
\cite{SpaRSA} proposed a framework, \sparsa, for solving the general problem
\begin{equation}
\min_{\coefv\in\reals^{\natoms}} f(\coefv) + \lambda\reg(\coefv).
\label{eq:general.problem}
\end{equation}
be a smooth and convex function, while $\reg:\reals^{\natoms} \rightarrow
\reals$ only needs to be finite and convex in $\reals^{\natoms}$.  This
formulation, which is a particular case of the Proximal Method framework
developed by Nesterov~\cite{nesterov07}, includes as important particular cases
the Lasso, Group-Lasso and HiLasso problems by setting $f(\cdot)$ as the
reconstruction error and then choosing the corresponding regularizers for
$\reg(\cdot)$.  When the regularizer, $\reg(\cdot)$, is group separable, the
optimization can be subdivided into smaller problems, one per group. The
framework becomes powerful when these sub-problems can be solved
efficiently. This is the case of the Lasso and Group Lasso (with non
overlapping groups) settings, and also of the HiLasso, as we will show later
in this Section. In all cases, the solution of the sub-problems are obtained
in linear time.

The \sparsa\ algorithm generates a sequence of iterates $\{ \coefv\iter{t}
\}_{t \in \naturals}$ that, under certain conditions, converges to the
solution of \eqref{eq:general.problem}.  At each iteration,
$\coefv\iter{t+1}$ is obtained by solving
\begin{equation}
\min_{\mat{z}\in\reals^{\natoms}}\;(\mat{z}-\coefv\iter{t})\transp \nabla f(\coefv\iter{t}) + \frac{\alpha\iter{t}}{2} \norm{ \mat{z} - \coefv\iter{t}}^2_2 + \lambda \reg(\mat{z}),
\label{eq:sparsa-subproblem}
\end{equation}
\noindent for a sequence of parameters $\{ \alpha\iter{t} \}_{t \in
  \naturals}$, $\alpha\iter{t}=\alpha_0\eta^t$, where $\alpha_0>0$ and
$\eta>1$ need to be chosen properly for the algorithm to converge (see
\cite{SpaRSA} for details). It is easy to show that
(\ref{eq:sparsa-subproblem}) is equivalent to
\begin{equation}
\min_{\mat{z}\in\reals^{\natoms}} \frac12 \norm{\mat{z} - \mat{u}\iter{t}}^2_2 + \frac{\lambda}{\alpha\iter{t}} \reg(\mat{z}),
\label{eq:sparsa-subproblemEq}
\end{equation}
\noindent where $ \mat{u}\iter{t}= \coefv\iter{t} - \frac{1}{\alpha\iter{t}}\nabla f(\coefv\iter{t}).$
In this new formulation, it is clear that the first term in the
cost function can be separated element-wise. Thus, when the
regularization function $\reg(\mat{z})$ is group separable, so is
the overall optimization, and one can solve
(\ref{eq:sparsa-subproblemEq}) independently for each group,
leading to
\[
\coefv\sblock{G}\iter{t+1} = \arg\min_{\mat{z}\in\reals^{|\group|}} \frac12 \norm{ \mat{z} -
\mat{u}\iter{t}\sblock{G}}^2_2 + \frac{\lambda}{\alpha\iter{t}}
\regG(\mat{z}),
\]
$\mat{z}\sblock{G}$ being the corresponding variable for the group. In the case of HiLasso, this becomes,
\begin{equation}
\coefv\sblock{G}\iter{t+1} = \arg\min_{\mat{z}\in\reals^{|\group|}}
\frac{1}{2}\norm{\mat{z}-\mat{w}}_2^2 +
\frac{\lambda_2}{\alpha\iter{t}} \norm{\mat{z}}_2 + \frac{\lambda_1}{\alpha\iter{t}}\norm{\mat{z}}_1,
\label{eq:sparsa:sub}
\end{equation}
\noindent where we have defined $\mat{w} = \mat{u}\iter{t}\sblock{G}$.
Problem (\ref{eq:sparsa:sub}) is a second order cone program
(\textsc{socp}), for which one could use generic solvers. However, since it
needs to be solved many times within the \sparsa\ iterations, it is crucial
to solve it efficiently. %
It turns out that \refeq{eq:sparsa:sub}
  admits a closed form solution with cost linear in the dimension of
  $\mat{w}$. By inspecting the subgradient of \refeq{eq:sparsa:sub} for the
  case where the optimum $\opt{\mat{z}} \neq 0$,
\[
\mat{w} - \left(1+\frac{\tilde{\lambda}_2}{\norm{\opt{\mat{z}}}_2}\right)\opt{\mat{z}} \in \tilde{\lambda}_1\partial\norm{\opt{\mat{z}}}_1,
\]
where we have defined $\tilde{\lambda}_2 = \lambda_2/\alpha\iter{t}$ and
$\tilde{\lambda}_1 = \lambda_1/\alpha\iter{t}$. If we now define
$C(\opt{\mat{z}}) = 1+\tilde{\lambda}_2/\norm{\opt{\mat{z}}}_2$, we observe
that each element of $C(\opt{\mat{z}})\opt{\mat{z}}$ is the solution of the well
known scalar soft thresholding operator, 
\begin{equation}
\opt{z}_i = \frac{1}{C(\opt{\mat{z}})}\sgn(w_i)\max \{0,|w_i|-\tilde{\lambda}_1\}=\frac{h_i}{C(\opt{\mat{z}})}\,, \quad i=1,\ldots,\groupsize,
\label{eq:sparsa:sub:2}
\end{equation}
where we have defined $h_i = \sgn(w_i)\max \{0,|w_i|-\tilde{\lambda}_1\}$,
the result of the scalar thresholding of $\mat{w}$. Taking squares on both
sides of \refeq{eq:sparsa:sub:2} and summing over $i=1,\ldots,\groupsize$ we
obtain
\[
\norm{\opt{\mat{z}}}_2^2 = C^2(\opt{\mat{z}})\norm{\mat{h}}_2^2 =
 \frac{\norm{\opt{\mat{z}}}_2^2}{(\norm{\opt{\mat{z}}}_2 + \tilde{\lambda}_2)^2}\norm{\mat{h}}_2^2,
\]
from which the equality $\norm{\opt{\mat{z}}}_2 = \norm{\opt{\mat{h}}}_2 -
\tilde{\lambda}_2$ follows. Since all terms are positive, this can only hold
as long as $\norm{\opt{\mat{h}}}_2 > \tilde{\lambda}_2$, which gives us a
vector thresholding condition on the solution $\opt{\mat{z}}$ in terms of
$\norm{\mat{h}}_2$. It is easy to show that $\norm{\opt{\mat{h}}}_2 \leq \tilde{\lambda}_2$ is a sufficient condition for $\opt{\mat{z}}= 0$. Thus we obtain,
\begin{equation}
\coefv\sblock{G}\iter{t+1} = \left\{
\begin{array}{ccc}
\frac{\max\{0,\norm{\mat{h}}_2-\tilde{\lambda}_2\}}{\norm{\mat{h}}_2}\mat{h}&,&\norm{\mat{h}}_2 > 0 \\
\mat{0}&,& \norm{\mat{h}}_2 = 0.
\end{array}
\right.
\label{eq:sparsa:sub:sol}
\end{equation}
The above expression requires $\groupsize$ scalar thresholding operations,
and one vector thresholding, which is also linear with respecto to the group
size $\groupsize$. Therefore, for all groups, the cost of solving the
subproblem \refeq{eq:sparsa:sub} is linear in $\ndims$, the same as for
Lasso and Group Lasso.  The complete HiLasso optimization algorithm is
summarized in Algorithm~\ref{alg1}. The parameter $\eta$ has very little
influence in the overall performance (see \cite{SpaRSA} for details); we
used $\eta=2$ in all our experiments.  Note that, as expected, the solution
to the sub-problem for the cases $\lambda_2=0$ or $\lambda_1=0$, corresponds
respectively to scalar soft thresholding and vector soft thresholding. In
particular, when $\lambda_2=0$, the proposed optimization reduces to the
Iterative Soft Thresholding algorithm \cite{daubechies04}.

\begin{algorithm}[t]
{\footnotesize
\caption{{\footnotesize HiLasso optimization algorithm.}}
\label{alg1}
\SetKw{Init}{Initialize}
\SetKw{Set}{set}
\SetKw{Choose}{choose}
\SetCommentSty{textit}
\KwIn{Data $\datam$, dictionary $\dictm$, group set $\groupset$, constants $\alpha_0>0$, $\eta>1$, $c>0$, 
$0<\alpha_{\textrm{min}}<\alpha_{\textrm{max}}$}
\KwOut{The optimal point $\opt{\coefv}$}
\Init $t := 0, \coefv\iter{0} := \mat{0}$\;
\While{stopping criterion is not satisfied}{
\Choose $\alpha\iter{t}\in [\alpha_{\textrm{min}},\alpha_{\textrm{max}}]$\;
\Set $\mat{u}\iter{t} :=  \coefv\iter{t} - \frac{1}{\alpha\iter{t}}\nabla f(\coefv\iter{t})$\;
\While{stopping criterion is not satisfied}
{%
\tcp{Here we use the group separability of \eqref{eq:sparsa-subproblemEq} and solve \eqref{eq:sparsa:sub} for each group}
\For{$i := 1$ to $\ngroups$}{
Compute $\coefv\iter{t+1}\sblock{G}$ as the solution to \refeq{eq:sparsa:sub:sol}\;
}
\Set $\alpha\iter{t+1} := \eta\alpha\iter{t}$\;
}
\Set $t := t + 1$ \;
}}
\end{algorithm}


\subsection{Optimization of the Collaborative HiLasso}
\label{sec:collaborative-solution}
The multi-signal (collaborative) case is equivalent to the
  one-dimensional case where the signal is a concatenation of the columns of
  $\datam$, and the dictionary is an
  $\nsamples\ndims{\times}\nsamples\natoms$ block-diagonal matrix, where each
  of the $\nsamples$ blocks is a copy of the original dictionary
  $\dictm$. However, in practice, it is not needed to build such (possibly
  very large) dictionary, and we can operate directly with the matrices
  $\dictm$ and $\datam$ to find $\coefm$.  If we define the matrix
  $\mat{U}\iter{t} \in \reals^{\ndims{\times}\nsamples}$ whose $i$-th column
  is given by $ \mat{u}_i\iter{t}= \coefv_i\iter{t} -
  \frac{1}{\alpha\iter{t}}\nabla f(\coefv_i\iter{t})$, we get the following
  SpaRSA iterates,
\[
\coefm\iter{t+1} = \arg\min_{\mat{Z} \in \reals^{\ndims{\times}\nsamples}} \frac{1}{2}\norm{\mat{Z}-\mat{U}\iter{t}}_F^2 + \frac{\lambda_2}{\alpha\iter{t}} \norm{\mat{Z}}_F + \frac{\lambda_1}{\alpha\iter{t}}  \sum_{j=1}^{\nsamples}  \norm{\mat{z}_j}_1,
\]
which again is group separable, so that it can be solved as $\ngroups$
independent problems in the corresponding bands of $\mat{U}\iter{t}$,
\[
(\coefm\iter{t+1})^G = \arg\min_{\mat{Z} \in \reals^{\groupsize{\times}\nsamples}} \frac{1}{2}\norm{\mat{Z}-(\mat{U}\iter{t})^G}_F^2 + \frac{\lambda_2}{\alpha\iter{t}} \norm{\mat{Z}}_F + \frac{\lambda_1}{\alpha\iter{t}} \sum_{j=1}^{\nsamples}  \norm{\mat{z}_j}_1.
\]
The correspondent closed form solutions for these subproblems, which are
obtained in an analogous way to
\refeq{eq:sparsa:sub:2}--\refeq{eq:sparsa:sub:sol}, are given by
\begin{equation}
(\coefm\iter{t+1})^G  = \left\{
\begin{array}{ccc}
\frac{\max\{0,\norm{\mat{H}}_F-\tilde{\lambda}_2\}}{\norm{\mat{H}}_F}\mat{H}&,&\norm{\mat{H}}_F > 0 \\
\mat{0}&,& \norm{\mat{H}}_F = 0
\end{array}
\right.,\quad h_{ij} = \sgn(w_{ij})\max\{0,|w_{ij}|-\tilde{\lambda}_1\},
\label{eq:collaborative-sparsa:sub:sol}
\end{equation}
and we have defined $\mat{W} := (\mat{U}\iter{t})^G$. 

As mentioned in Section~\ref{sec:background:literature}, \cite{JenattonBach}
addresses a wide spectrum of hierarchical sparse models for coding and
dictionary learning. They propose a proximal method optimization procedure
that, when restricted to the formulation of HiLasso, is very similar to the
one developed in Section~\ref{sec:opt.single}. The main difference with our
method is that they solve the sub-problem \eqref{eq:sparsa-subproblemEq}
using a dual approach (based on conic duality) that finds the exact solution
in a finite number of operations. Our method, being tailored to the specific
case of HiLasso, provides such solution in closed form, requiring just two
thresholdings, both linear in the dimension of $\datam$, $\nsamples{\times}\ndims$.
\label{obs:optimization1}


\section{Theoretical guarantees}
\label{sec:theory}

In our current theoretical analysis, we study the case of a single
measurement vector (signal) $\datav$ (we comment on the collaborative case
at the end of this section), and assume that there is no measurement noise
or perturbation, so that $\datav=\dictm\coefv$. Without loss of generality,
we further assume that the cardinality $|G_r|=g, r=1,\ldots,\ngroups$, that
is, all groups in $\groupset$ have the same size.  The goal is to recover
the code $\coefv$, from the observed $\datav$, by solving the noise-free
HiLasso problem:
\begin{equation}
\label{eq:hilasso2} 
\min_{\coefv\in\reals^{\natoms}} 
\blc \lambda \regG(\coefv) + (1-\lambda)\norm{\coefv}_1 
\subjto 
\datav=\dictm\coefv \brc.
\end{equation}
Note that we have replaced the two regularization parameters $\lambda_1$ and
$\lambda_2$ by a single parameter $\lambda$, since scaling does not effect
the optimal solution. Therefore, we can always assume that
$\lambda_1+\lambda_2=1$.

Our goal is to develop conditions under which the HiLasso program of
(\ref{eq:hilasso2}) will recover the true unknown vector $\coefv$.  
As we will see, the resulting set of recoverable signals is a superset
of those recoverable by Lasso, that is, HiLasso is able to recover signals
for which Lasso (or Group Lasso) will fail to do so.

We assume throughout this section that $\coefv$ has group sparsity $k$,
namely, no more than $k$ of the group vectors $\coefv\sblock{\group_i},
i=1,\ldots,\ngroups$, have non-zero norm. In addition, within each group, we
assume that not more than $s$ elements are non zero, that is,
$\|\coefv\sblock{\group}\|_0 \leq s$.

For $\lambda=1$, (\ref{eq:hilasso2}) reduces to the Group Lasso
problem, \refeq{eq:group-lasso}, whereas with $\lambda=0$,
(\ref{eq:hilasso2}) becomes equivalent to the Lasso problem,
\refeq{eq:lasso}. Both cases have been treated previously in the literature
and sufficient conditions have been derived on the sparsity levels and on
the dictionary $\dictm$ to ensure that the resulting optimization problem
recovers the true unknown vector $\coefv$.  For example, in
\cite{EM09,CRT06,C08}, conditions are given in terms of the restricted
isometry property (RIP) of $\dictm$. In an alternative line of work,
recovery conditions are based on coherence measures, which are easier to
compute \cite{EKB09,tropp04}. Here, we follow the same spirit and consider
coherence bounds that ensure recovery using the HiLasso approach. We also
draw from \cite{ER10} to briefly describe conditions under which the
probability of error of recovering the correct groups, using the special
case of the C-HiLasso with $\lambda_1=0$ (C-GLasso), falls exponentially to
$0$ as the number of collaborating samples $\nsamples$ grows.  Finally,
recent theoretical results on block sparsity were reported in
\cite{stojnic09}.\label{obs:stojnic} In particular, bounds on the number of measurements
required for block sparse recovery were developed under the assumption that
the measurement matrix $\dictm$ has a basis of the null-space distributed
uniformly in the Grassmanian. The model is a block-sparse model, without
hierarchical or collaborative components.

In this section we extend the group-wise indexing notation to refer both to
subsets of rows and columns of arbitrary matrices as $\mat{W}\sblock{F,G} :=
\{w_{ij}: i \in F, j \in G\}$.  This is, $\mat{W}\sblock{F,G}=
\mat{I}\sblock{F}^T\mat{W}\mat{J}\sblock{G}$, where $\mat{I}$ and $\mat{J}$
are the identity matrices of the column and row spaces of $\mat{W}$
respectively. We define the sets $\Omega=\setdef{1,2,\ldots,\natoms}$ and
$\Gamma=\setdef{1,2,\ldots,g}$, and use $\cpl{S}$ to denote the complement
of a set of indices $S$, either with respect to $\Omega$ or $\Gamma$,
depending on the context. The set difference between $S$ and $T$ is denoted
as $S \setminus T$, $\emptyset$ represents the empty set, and $|S|$ denotes
the cardinality of $S$.

\subsection{Block-Sparse Coherence Measures}
\label{sec:theory:block-coherence}
We begin by reviewing previously proposed coherence measures. For a given
dictionary $\dictm$, the (standard) coherence is defined as
$\mutualco\defeq\max_{i, j \neq i \in \Gamma} |\dictv_{i}\transp
\dictv_{j}|$. This coherence was extended to the block-sparse setting in
\cite{EKB09}, leading to the definition of \emph{block coherence}:
\[
\blockco\defeq\max \left\{
\frac{1}{g}\rho(\dictm\sblock{G}\transp\dictm\sblock{F}),\;
G,F  \in \groupset, G \neq F
\right\},
\]
where $\rho(\cdot)$ is the spectral norm, that is,
$\rho(\auxm)\defeq\lambda^{1/2}_{\max}(\auxm\transp\auxm)$, with
$\lambda_{\max}(\bbw)$ denoting the largest eigenvalue of the positive
semi-definite matrix $\bbw$. An alternate atom-wise measure of block
coherence is given by the \emph{cross coherence},
\begin{equation}\label{eq:crossco}
\crossco\defeq\max \left\{ \max
  \left\{|\dictv_i^T\dictv_j|, \; i \in G, j \in F \right\} 
G,F  \in \groupset, G \neq F
\right\}.
\end{equation}
When $g=1$ (each block is a singleton), $\dictm\sblock{G_r}=\dictv_r$, so
that as expected, $\crossco=\blockco=\mutualco$. %
While $\blockco$ and $\crossco$ quantify global properties of the dictionary
$\dictm$, local block properties are characterized by the
\emph{sub-coherence}, defined as
\begin{align}\label{eq:subco}
 \subco \defeq \max \left\{ \max \left\{|\dictv_i^T\dictv_j|,
\;  i,j \in G, i \neq j \right\}\; G \in \groupset \right\}.
\end{align}
We define $\subco=0$ for $g=1$. Clearly, if the columns of
$\dictm\sblock{\group}$ are orthonormal for each group $G$, then
$\subco=0$. Assuming the columns of $\dictm$ have unit norm, it can be
easily shown that $\mutualco$, $\subco$, $\crossco$ and $\blockco$ all lie
in the range $[0,1]$. In addition, we can easily prove that
$\subco,\blockco,\crossco \leq \mutualco$.  In our setting, $\coefv$ is
block sparse, but has further internal structure: each sub-vector of
$\coefv$ is also sparse. In order to quantify our ability to recover such
signals, we expect that an appropriate coherence measure will be based on
the definition of block sparsity, but will further incorporate the internal
sparsity as well. Let $\mat{M}\defeq\dictm\transp\dictm$ denote the Gram matrix
of $\dictm$. Then, the standard block coherence $\blockco$ is defined in
terms of the largest singular value of an off-diagonal $g{\times}g$
sub-block of $\mat{M}$. In a similar fashion, we will define \emph{
  sparse block coherence} measures in terms of \emph{sparse singular
  values}. As we will see, two different definitions will play a role,
depending on where exactly the sparsity within the block enters. To define
these, we note that the \emph{spectral norm} $\rho(\auxm)$ of a matrix
$\auxm$ can be defined as
$$
\rho(\auxm)\defeq\max_{\mat{u},\mat{v}}|\mat{u}\transp \auxm\mat{v}|\quad \subjto
\norm{\mat{u}}_2=1,\norm{\mat{v}}_2=1.
$$
Alternatively, we can define $\rho(\auxm)$ as the largest singular value of
$\auxm$,
$\rho(\auxm)\defeq\sigma_{\max}(\auxm)=\sqrt{\lambda_{\max}(\auxm\transp\auxm)},$
$$
\lambda_{\max}(\auxm\transp\auxm)\defeq\max_{\mat{v}}\mat{v}\transp(\auxm\transp\auxm)\mat{v} \quad
\subjto\;\norm{\mat{v}}_2=1.
$$
We now develop sparse analogs of $\rho(\auxm)$ and
$\lambda_{\max}(\auxm\transp\auxm)$. As we will see, the simple square-root
relation no longer holds in this case. The \emph{largest sparse singular
  value} is defined as \cite{AEJL04}:
\begin{equation}
\label{eq:rmaxs} \rho^{ss}(\auxm)\defeq\max_{\mat{u},\mat{v}}|\mat{u}\transp \auxm\mat{v}|
\quad \subjto \norm{\mat{u}}_2=1,\norm{\mat{v}}_2=1,\norm{\mat{u}}_0 \leq s,\norm{\mat{v}}_0 \leq s.
\end{equation}
Similarly, the \emph{largest sparse eigenvalue} of $\auxm\transp\auxm$ is
defined as \cite{AEJL04,ZHT03,MWA06},
\begin{equation}
\label{eq:lmaxs} 
\lambda_{\max}^s(\auxm\transp\auxm) \defeq \max_{\mat{v}}\mat{v}\transp (\auxm\transp\auxm) \mat{v}
\quad \subjto \norm{\mat{v}}_2=1,\norm{\mat{v}}_0 \leq s.
\end{equation}
The \emph{sparse matrix norm} is then given by
\begin{equation}
\label{eq:norms} 
\rho^s(\auxm)\defeq\sqrt{\lambda_{\max}^s(\auxm\transp\auxm)}.
\end{equation}
\label{obs:sparse-pca} Note that, in general, $\rho^s(\auxm)$ is not equal
to $\rho^{ss}(\auxm)$. It is easy to see that $\rho^{ss}(\auxm) \leq
\rho^s(\auxm)$. For any matrix $\auxm$,
$\rho^{ss}(\auxm)=\rho(\auxm\sblock{F,G})$ and
$\rho^{s}(\auxm)=\rho(\auxm\sblock{T})$, where $F,G,T$ are subsets of
$\Gamma=\setdef{1,2,\ldots,\groupsize}$ of size $s$, chosen to maximize the
corresponding singular value. Using (\ref{eq:rmaxs}) and (\ref{eq:norms}),
we define two sparse block coherence measures:
\begin{eqnarray}
\label{eq:ssbc} \blockco^{ss} &\defeq& \max 
\left\{\frac{1}{g}
\rho^{ss}(\dictm\sblock{G}\transp\dictm\sblock{F}),
\, G,F \in \groupset, G \neq F
\right\},\\
\label{eq:sbc} \blockco^s&\defeq&\max \left\{ 
\frac{1}{g} \rho^s(\dictm\sblock{G}\transp\dictm\sblock{F}),
\, G,F \in \groupset, G \neq F
\right\}.
\end{eqnarray}
The choice of scaling is to ensure that
$\blockco^s,\blockco^{ss} \leq \blockco.$

Note that, while $\rho^s(\auxm)$ (also referred to in the literature as
\emph{sparse principal component analysis} (SPCA)) and $\rho^{ss}(\auxm)$
are in general NP-hard to compute, in many cases they can be computed
exactly, or approximated, using convex programming
techniques~\cite{AEJL04,ZHT03,MWA06}.

The following proposition establishes some relations between these
new definitions and the standard coherence measures.
\begin{proposition}
\label{prop:sbc}  The sparse block-coherence measures
$\blockco^{ss},\blockco^s$ satisfy
\begin{equation}
0 \leq \blockco^{ss} \leq \frac{s}{g}\mutualco, \quad 0 \leq \blockco^s \leq
\sqrt{\frac{s}{g}}\mutualco.
\end{equation}

\end{proposition}
\begin{proofi} The inequalities
  $\blockco^{ss},\blockco^{s} \geq 0$ follow immediately
  from the definition. We obtain the upper bounds by rewriting
  $\rho^{ss}(\auxm)$ and $\rho^{s}(\auxm)$ and then using the Ger\v{s}gorin
  Theorem,
\begin{eqnarray}
\rho^{ss}(\auxm) & = &\lambda_{\max}^{1/2}(\auxm\sblock{F,G}\transp\auxm\sblock{F,G}) 
\stackrel{(a)}{\leq} \sqrt{\max_l \sum_{r=1}^{s}|e_{lr}|}
\leq \sqrt{s \max_{l,r} |e_{lr}|}, \label{eq:rhoss-bound} \\
\rho^s(\auxm) & = & \lambda_{\max}^{1/2}(\auxm\sblock{T}\transp\auxm\sblock{T}) 
\stackrel{(b)}{\leq} \sqrt{\max_l \sum_{r=1}^{s}|e'_{lr}|} \leq 
\sqrt{s \max_{l,r} |e'_{lr}|}, \label{eq:rhos-bound}
\end{eqnarray}
where $e_{lr}$ and $e'_{lr}$ are the elements of
$\mat{E}=\auxm\sblock{F,\Gamma}\transp\auxm\sblock{F,\Gamma}$ and
$\mat{E}'=\auxm\transp\auxm$, and $(a)$, $(b)$ are a consequence of
Ger\v{s}gorin's disc theorem.

The entries of $\auxm=\dictm\sblock{G_i}\transp\dictm\sblock{G_j}$ for $i
\neq j$ have absolute value smaller than or equal to $\mutualco$, and the
size of $\auxm$ is $g \times g$. Therefore, $|e_{k \ell}| \leq s\mutualco^2$
and $|e'_{k \ell}| \leq \groupsize\mutualco^2$. Substituting these values
into (\ref{eq:rhoss-bound}) and (\ref{eq:rhos-bound}) concludes the proof of
the upper bounds on $\blockco^{ss}$ and $\blockco^s$.\end{proofi}

\subsection{Recovery Proof}
%
%
%
\def\dictmGo{\dictm\sblock{G_0}} 
\def\dictmGoc{\dictm\sblock{\cpl{G_0}}} 
\def\dictmSo{\dictm\sblock{S_0}}
\def\dictmTo{\dictm\sblock{T_0}} 
\def\dictmSoc{\dictm\sblock{\cpl{S_0}}}
\def\coefvGo{\coefv\sblock{G_0}} 
\def\coefvGoc{\coefv\sblock{\cpl{G_0}}} 
\def\coefvSo{\coefv\sblock{S_0}}
\def\coefvTo{\coefv\sblock{T_0}} 
\def\coefvSoc{\coefv\sblock{\cpl{S_0}}}
\def\dictmGop{\dictm\sblock{G_0'}} 
\def\dictmSop{\dictm\sblock{S_0'}}
\def\dictmBp{\dictm\sblock{B}} 
\def\dictmRp{\dictm\sblock{\cpl{B}}}
\def\coefvGop{\coefv'\sblock{G_0'}} 
\def\coefvSop{\coefv'\sblock{S_0'}}
\def\coefvBp{\coefv'\sblock{B}} 
\def\coefvRp{\coefv'\sblock{\cpl{B}}}
\def\rhocSoGoc{\rhoc{\mathcal{S}_0,\cpl{\mathcal{G}}_0}}
\def\rhocSoSo{\rhoc{\mathcal{S}_0,\mathcal{S}_0}}

\begin{figure}
\centering
\includegraphics[width=0.45\textwidth]{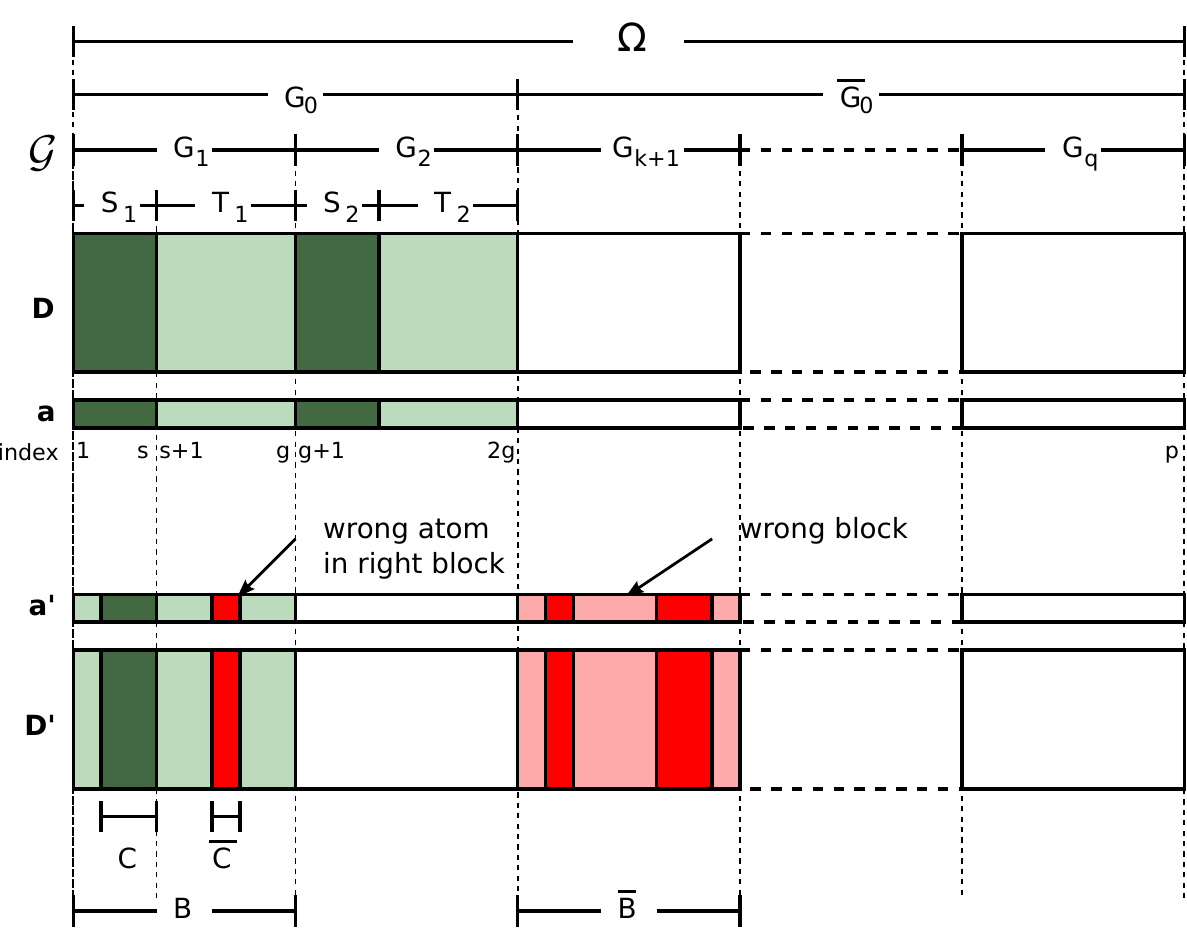}\hspace{3ex}%
\includegraphics[width=0.375\textwidth]{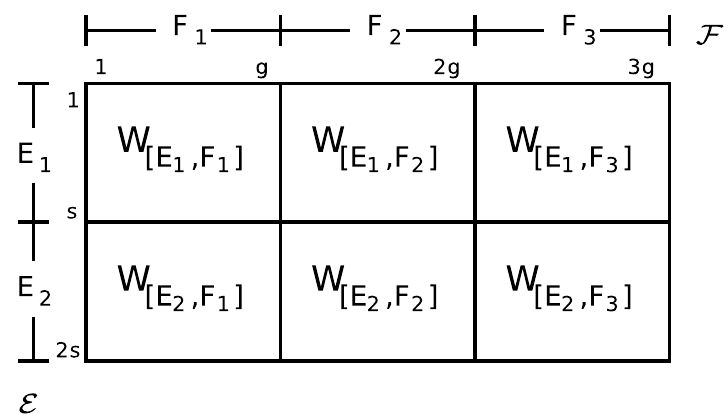}
\caption{\label{fig:indexing}Left: Indexing conventions, here shown for
  $g=8$, $k=2$ and $s=3$. Shaded regions correspond to active
  elements/atoms. Active blocks are light-colored, and active
  elements/coefficients are dark colored. Here $\coefv'$ represents an
  alternate representation of $\datav$, $\datav=\dictm\coefv'$. Blocks and
  atoms that are not part of the true solution $\coefv$ are marked in
  red. Right: partitioning of a matrix $\mat{W}$ performed by the measure
  $\rhoc{\mathcal{E},\mathcal{F}}(\mat{W})$ with $\mathcal{E}=\{E_1,E_2\}$
  and $\mathcal{F}=\{F_1,F_2,F_3\}$, where $|E_i|=s$ and $|F_j|=g$. }
\end{figure}

Our main recovery result is stated as follows. Suppose that $\coefv$ is a
block $k$-sparse vector with blocks of length $g$, where each block has
sparsity exactly $s$,\footnote{These conditions are non-limiting, since we
  can always complete the vector with zeros.} and let
$\datav=\dictm\coefv$. We rearrange the columns in $\dictm$ and the
coefficients in $\coefv$ so that the first $k$ groups,
$\{G_1,G_2,\ldots,G_k\}$ correspond to the non-zero (active) blocks.  Within
each block $G_i, i\leq k$, the first $s$ indices, represented by the set
$S_i$, correspond to the $s$ nonzero coefficients in that block, and the
index set $T_i=G_i \setminus S_i$ represents its $(g-s)$ inactive elements,
so that $G_i=[S_i\;T_i]$. The set $G_0=\bigcup_{i=1}^{k}G_i$ contains the
indices of all the active blocks of $\coefv$, whereas $\cpl{G_0}=\Omega
\setminus G$ contains the inactive ones. Similarly,
$S_0=\bigcup_{i=1}^{k}S_i$ contains the indices of all the active
coefficients/atoms in $\coefv$ and $\dictm$ respectively, $\cpl{S_0}=\Omega
\setminus S_0$ indexes the inactive coefficients/atoms in $\coefv$/$\dictm$,
and $T_0=\bigcup_{i=1}^{k}T_i$ indexes the inactive coefficients/atoms
within the active blocks.  These indexing conventions are exemplified in
Figure~\ref{fig:indexing}(left). With these conventions we can write
$\datav=\dictmGo\coefvGo=\dictmSo\coefvSo.$

An important assumption that we will rely on throughout, is that the
columns of $\dictmGo$ must be linearly independent for any $G_0$ as defined
above.
Under this assumption, $\dictmSo\transp\dictmSo$ is invertible and we can
define the pseudo-inverse $\pinv \defeq
(\dictmSo\transp\dictmSo)^{-1}\dictmSo\transp.$ For reasons that will become
clear later, we will also need a second, oblique pseudo-inverse, $\oinv
\defeq
(\dictmSo\transp(\mat{I}-\mat{P})\dictmSo)^{-1}\dictmSo\transp(\mat{I}-\mat{P})$,
where $\mat{P}$ is an orthogonal projection onto the range of $\dictmTo$,
that is, $\mat{P}\dictmTo=\dictmTo$. It is easy to check that
\begin{equation}
\label{eq:oblique-properties}
\oinv\dictmTo=\mat{0},\quad\mathrm{and}\quad \oinv\dictmSo=\mat{I}.
\end{equation}
Equipped with these definitions we can now state our main result.

\begin{theorem}
  \label{thm:sc} Let $\coefv$ be a block $k$-sparse vector with blocks of
  length $g$, where each block has sparsity $s$. Let $\datav=\dictm\coefv$
  for a given matrix $\dictm$. A sufficient condition for the HiLasso
  algorithm (\ref{eq:hilasso2}) to recover $\coefv$ from $\datav$ is that,
  for some $\alpha \leq 1$,
\begin{eqnarray}
\label{eq:sc1}
\rhocSoGoc(\oinv \dictmGoc)   & < & \alpha, \\
\label{eq:sc3}
\|\pinv \dictmGoc\|_{1,1} & < & \gamma,\quad \gamma \leq 
1 + \frac{\lambda(1-\alpha)}{ \sqrt{g}(1-\lambda) },\\
\label{eq:sc4}  
\|\pinv\dictmTo\|_{1,1} & < & 1.
\end{eqnarray}
Here $\rhoc{\mathcal{E},\mathcal{F}}(\auxm) := \max_{F \in \mathcal{F}}
\sum_{E \in \mathcal{E}} \rho(\auxm\sblock{E,F})$, is the block spectral
norm defined in \cite{EKB09}, the blocks defined by the sets of index sets
$\mathcal{E}$ and $\mathcal{F}$ (see Figure~\ref{fig:indexing}(right)). We also
define $\mathcal{S}_0=\setdef{S_i:i=1,\ldots,k}$,
$\cpl{\mathcal{G}}_0=\setdef{G_i:i=k+1,\ldots,\ngroups}$ and
$\mathcal{T}_0=\setdef{T_i:i=1,\ldots,k}$. Finally, $\norm{\auxm}_{1,1} :=
\max_{r} \norm{\auxv_r}_1$, where $\auxv_r$ is the $r$-th column of $\auxm$.
\end{theorem}
\label{obs:hilasso-vs-lasso}The above theorem can be interpreted as
follows. With $\gamma=1$, the conditions \refeq{eq:sc3}-\refeq{eq:sc4} are
sufficient both for Lasso ($\lambda=0$) and HiLasso to recover
$\coefv$. However, if there exists a $\gamma > 1$ for which condition
\refeq{eq:sc3} holds, then HiLasso will be able to recover $\coefv$ in a
situation where Lasso is not guaranteed to do so. The idea is that, for $0 <
\lambda < 1$, HiLasso trades off between the minimization of its \cost{1}
and \cost{2} terms, by tightening the \cost{2} term ($\alpha \leq 1$) to
improve group recovery, while loosening the \cost{1} term ($\gamma >
1$). 
Also, although not yet clear from conditions
  \refeq{eq:sc1}--\refeq{eq:sc4}, we will see in Theorem~\ref{thm:mu} that
  the final data independent bounds are also a relaxation of the ones
  corresponding to Group Lasso when the solutions are block-dense.
Therefore, the proposed model outperforms both standard Lasso and Group
Lasso with regard to recovery guarantees. This is also reflected in the
experimental results presented in the next section.

The sufficient conditions (\ref{eq:sc1})--(\ref{eq:sc4}) depend on
$\dictmSo$ and therefore on the nonzero blocks in $\coefv$, $G_0$, and the
nonzero locations within the blocks, $S_0$, which, of course, are not known
in advance.  Nonetheless, Theorem~\ref{thm:mu} provides sufficient
conditions ensuring that \refeq{eq:sc1}--\refeq{eq:sc4} hold, which are
independent of the unknown signals, and depend only on the dictionary
$\dictm$.


We now prove Theorem~\ref{thm:sc}.

\begin{proofi}
  To prove that (\ref{eq:hilasso2}) recovers the correct vector $\coefv$,
  let $\coefv'$ be an alternative solution satisfying
  $\datav=\dictm\coefv'$. We will show that
  $\lambda\regG(\coefv)+(1-\lambda)\norm{\coefv}_1 <
  \lambda\regG(\coefv')+(1-\lambda)\norm{\coefv'}_1$. Let the set
  $G_0$ contain the indices of all elements in the active blocks of
  $\coefv$. Let $G_0'$ contain the indices of the active blocks in
  $\coefv'$. Then $\datav=\dictmGo\coefvGo=\dictmGop\coefvGop$.

  By our assumptions, in each block of $\coefvGo$ there are exactly $s$
  nonzero values. Let the set $S_0 \subset G_0$ contain the indices of all
  nonzero elements in $\coefv$. We thus have $|S_0|=ks$. Using
  \refeq{eq:oblique-properties} we can write
\begin{equation}
\label{eq:oinv-identity} \coefvSo=\oinv \dictmSo\coefvSo=\oinv
\dictmGo \coefvGo = \oinv\dictmGop\coefvGop.
\end{equation}

To proceed, we separate $G_0'$ into two parts: $B=G_0' \cap G_0$, and
$\cpl{B}=G_0'\setminus G_0$, so that $G_0'=[B \; \cpl{B}]$ and
$\dictmGop\coefvGop=\dictm\sblock{B}\coefvBp + \dictm\sblock{\cpl{B}}\coefvRp$.
We can now rewrite \refeq{eq:oinv-identity} as 
\begin{equation}
\label{eq:coefv-decomp}
\coefvSo = \oinv\dictmBp\coefvBp + \oinv\dictmRp\coefvRp,
\end{equation}
\noindent and use the triangle inequality to obtain
\begin{equation}
\label{eq:coineq} 
\regG(\coefvSo) \leq 
\regG(\oinv\dictmBp\coefvBp)+\regG(\oinv\dictmRp\coefvRp).
\end{equation}
We now analyze the two terms in the right hand side of (\ref{eq:coineq})
using \cite[Lemma~3]{EKB09}:
\begin{lemma}
  \label{lemma:1inq} Let $\mat{v} \in \reals^\natoms$ be a vector, $\auxm\in
  \reals^{\ndims{\times}\natoms}$ be a matrix, $\mathcal{F}$ be a partition
  of $\Omega=\setdef{1,2,\ldots,\natoms}$, and $\mathcal{E}$ a partition of
  $\setdef{1,\ldots,\ndims}$. We then have that, $\regG(\auxm\bv)
  \leq \rhoc{\mathcal{E},\mathcal{F}}(\auxm)
  \regG(\bv)$.\footnote{Note that the statement of
    Lemma~\ref{lemma:1inq} as shown here is actually a slight generalization
    of \cite[Lemma~3]{EKB09}, where the groups in the partitions need not
    have the same size.}
\end{lemma}

Since $\cpl{B} \subset \cpl{G_0}$, it follows from (\ref{eq:sc1}) that
$\rhoc{\mathcal{S}_0,\cpl{\mathcal{B}} }(\oinv\dictmRp)<\alpha$ (here
$\cpl{\mathcal{B}}$ is the set of the blocks that comprise $\cpl{B}$). 
To analyze $\rhoc{\mathcal{S}_0,\mathcal{B}}(\oinv\dictmBp)$, we use its
definition,
\begin{equation}
\rhoc{\mathcal{S}_0,\mathcal{B}}(\oinv\dictmBp) = 
\max_{F \in \mathcal{B}} 
  \sum_{E \in \mathcal{S}_0} \rho((\oinv\dictm)\sblock{E,F}) 
= \max_{F \in \mathcal{B}} 
  \sum_{S_j: j=1,\ldots,k} \rho((\oinv\dictm)\sblock{S_j,F}),\label{eq:rhoc-HB}
\end{equation}
and analyze each of its terms.  By definition of $\mathcal{B}$, each $F \in
\mathcal{B}$ corresponds to some $G_i = [S_i\;T_i]$ for some $i \leq k$.  We
can thus write $(\oinv\dictm)\sblock{S_j,F}=
[\,(\oinv\dictm)\sblock{S_j,S_i} \; (\oinv\dictm)\sblock{S_j,T_i}\,]$. Then,
by recalling that $\oinv\dictmTo=\mat{0}$ we see that
$(\oinv\dictm)\sblock{S_j,T_i}=\mat{0}$ for all $i,j$. Now, when $i=j$ we
have $(\oinv\dictm)\sblock{S_j,S_i}=\mat{I}$, thus
$\rho((\oinv\dictm)\sblock{S_j,F})=\rho([\,\mat{I}\;\mat{0}\,]) = 1$. When
$i\neq j$, $(\oinv\dictm)\sblock{S_j,S_i}=\mat{0}$, and 
$\rho((\oinv\dictm)\sblock{S_j,F}) = \rho([\,\mat{0}\;\mat{0}\,]) = 0$ in
that case. From \refeq{eq:rhoc-HB} we conclude that
$\rhoc{\mathcal{S}_0,\mathcal{B}}(\oinv\dictmBp) = 1$. Plugging into
\refeq{eq:coineq} leads to
\begin{equation}
\label{eq:ineq-l2} 
\regG(\coefv) < \regG(\coefvBp)+ \alpha\regG(\coefvRp).
\end{equation}

For the \cost{1} term, we follow the same path as \refeq{eq:oinv-identity}
and \refeq{eq:coefv-decomp}, now using the Moore-Penrose pseudo-inverse
$\pinv$ instead, yielding $\coefvSo = \pinv\dictmBp\coefvBp +
\pinv\dictmRp\coefvRp$, from which $\norm{\coefv}_1 \leq
\norm{\pinv\dictmBp\coefvBp}_1+\norm{\pinv\dictmRp\coefvRp}_1$
follows. Using the fact that $\norm{\mat{W}\mat{v}}_{1,1} \leq
\norm{\mat{W}}_{1,1}\norm{\mat{v}}_1$~\cite{tropp04}, we get $\norm{\coefv}_1 \leq
\norm{\pinv\dictmBp}_{1,1}\norm{\coefvBp}_1 +
\norm{\pinv\dictmRp}_{1,1}\norm{\coefvRp}_1$. Now, since $B \subset G_0$,
and $\norm{\pinv\dictmGo}_{1,1} = 1$, we have that $\norm{\pinv\dictmBp}_{1,1}
\leq 1$. Together with condition \refeq{eq:sc4} this yields,
\begin{equation}
\label{eq:ineq-l1} 
\norm{\coefv}_1 < \norm{\coefvBp}_1 + \gamma\norm{\coefvRp}_1.
\end{equation}

Combining \refeq{eq:ineq-l2} and \refeq{eq:ineq-l1} into the HiLasso cost
function we get
\begin{equation}
\label{eq:ineq-hilasso-1}
\lambda\regG(\coefv) + 
(1-\lambda)\norm{\coefv}_1 \; <\;
\lambda \left[\, \regG(\coefvBp) + \alpha\regG(\coefvRp) \,\right] 
+ 
(1-\lambda)\left[\, \norm{\coefvBp}_1 + \gamma\norm{\coefvRp}_1 \,\right].
\end{equation}
Now, to finish the proof, we need to bound the right hand side of
\refeq{eq:ineq-hilasso-1} by $\lambda\regG(\coefv') +
(1-\lambda)\norm{\coefv'}_1$, in order to show that the alternate $\coefv'$
is not a minimum of the HiLasso problem. For any $\gamma$ satisfying
\[
\gamma \leq 1 +
\frac{\lambda(1-\alpha) \reg_{\groupset }(\coefvRp)}
{ (1-\lambda)\norm{\coefvRp}_1 },
\]
we have,
\begin{equation}
\label{eq:ineq-hilasso-2}
\lambda[\,\regG(\coefvBp)+\alpha\regG(\coefvRp)\,] +  (1-\lambda)[\,\norm{\coefvBp}_1+\gamma\norm{\coefvRp}_1\,]
\leq 
\lambda\regG(\coefv') +  (1-\lambda)\norm{\coefv'}_1,
\end{equation}
where we have used the fact that
$\norm{\coefv'}_1=\norm{\coefv'\sblock{B}}_1+\norm{\coefv'\sblock{\cpl{B}}}_1$ and
$\regG(\coefv')=\regG(\coefv'\sblock{B})+\regG(\coefv'\sblock{\cpl{B}})$.
To obtain a signal independent relationship between $\gamma$ and
$\alpha$, we bound $\regG(\coefvRp)$ in terms of
$\norm{\coefvRp}_1$,
$$\norm{\coefvRp}_1  
= \sum_{i}\norm{ \coefv'\sblock{\cpl{R}_i} }_1 \leq \sum_{i}\sqrt{\groupsize}
\norm{\coefv'\sblock{\cpl{B}_i}}_2 = \sqrt{g}\reg_{\groupsize}(\coefvRp),$$
\noindent resulting in the condition
\[
\gamma \leq 
1+ \frac{\lambda(1-\alpha)}{ (1-\lambda)\sqrt{g} }  \leq
1+ \frac{\lambda(1-\alpha) \reg_{\groupset }(\coefvRp)}{ (1-\lambda)\norm{\coefvRp}_1},
\]
which completes the proof.\end{proofi}

We conclude that we can guarantee recovery for every choice of $\lambda$ as
long as (\ref{eq:sc1})--(\ref{eq:sc4}) are satisfied. Note that when
$\lambda=0$ (Lasso mode) we get $\gamma \leq 1$, and, as expected,
\refeq{eq:sc3}--\refeq{eq:sc4} reduce to the Lasso recovery condition.
Also, if $\alpha=1$ we have $\gamma\leq 1$, meaning that we must tighten the
constrains related to the \cost{2} part of the cost function in order to
relax the \cost{1} part. For $\gamma > 1$, the HiLasso conditions are a
relaxation of the Lasso conditions, thus allowing for more signals to be
correctly recovered.

%
\def\pdict{c}
\def\pdictv{\vec{\pdict}}
\def\pdictm{\mat{C}}
\def\mup{\mutualco_{P}} \def\nup{\subco_{P}}
\def\scalm{\Lambda}
Theorem~\ref{thm:mu} below provides signal independent replacements of the
conditions \refeq{eq:sc1}--\refeq{eq:sc4}. The signal independent bound for
\refeq{eq:sc1} derived here, depends on coherence measures between the
dictionary $\dictm$ and its image under the projection $\mat{I}-\mat{P}$,
$\pdictm=(\mat{I}-\mat{P})\dictm$. Since $\mat{P}$ depends on $S_0$,
$\pdictm$ itself is signal dependent. Thus, we need to maximize also over
all possible sets $S_0$. These are defined as follows,
 \begin{align}
 \nup &\defeq
\max\left\{
\max\left\{
\max\left\{
   \frac{\dictv_i\transp \pdictv_j}{(\dictv_i\transp \pdictv_i)^{1/2}(\dictv_j\transp\pdictv_j)^{1/2}},\, i , j \in G, i \neq j 
 \right\}
,\,G \in \groupset \right\}
,\, S_0 \right\}\label{eq:pco-first},\\
 \mup^s &\defeq
\max\left\{
\max\left\{
 \frac{1}{g}
 \rho^s(\dictm\transp\sblock{G}\pdictm\sblock{F}),\,
 G, F \in \groupset, G \neq F
 \right\},
\, S_0 \right\},\label{eq:pco-2}\\
 \mup^{ss} &\defeq
\max\left\{
\max\left\{
 \frac{1}{g}\rho^{ss}(\dictm\sblock{G}\transp\pdictm\sblock{F}),\,
 G, F \in \groupset, G \neq F
 \right\}
,\, S_0 \right\},\label{eq:pco-3}\\
\label{eq:pco-last}
\zeta &\defeq 
\max\left\{
\max \{  (\dictv_i\transp\pdictv_i)^{-1/2} : i=1,\ldots,\natoms 
\},\,S_0 \right\}.
 \end{align}
We are now in  position to state the theorem.

\begin{theorem}
  \label{thm:mu} Let $\crossco$, $\nu_P$, $\mup^{s}$, $\mup^{ss}$ and
  $\zeta$ be the coherence measures defined respectively in
  \refeq{eq:crossco} and \refeq{eq:pco-first}--\refeq{eq:pco-last}. Then the
  conditions (\ref{eq:sc1})--(\ref{eq:sc4}) are satisfied if
\begin{eqnarray}\label{eq:sc1s}
\frac{ \zeta^2 kg\mup^s } {1-(s-1)\nup+g\mup^{ss}  (k-1)\zeta^2 } & \leq & \alpha,\\
\label{eq:sc3s} 
\frac{ ks\crossco } {1- (s-1)\subco  - (k-1)s\crossco} &<& \gamma, \\
\label{eq:sc4s} 
\frac{ ks\subco } {1- (s-1)\subco - (k-1)s\crossco} &<& 1.
\end{eqnarray}
We also require the denominators in \refeq{eq:sc1s}--\refeq{eq:sc4s} to be
positive.  Note that, although the interpretation of \refeq{eq:sc1s} is
rather counter-intuitive, it is easy to check that $\mup^{s}, \mup^{ss} \leq
\blockco$. This can be seen when $s=g$ (a case included in our theorems), in
which case $\mat{P}=\mat{0}$, $\pdictm=\dictm$, and
$\mup^{s}=\mup^{ss}=\blockco$. Therefore, the condition \refeq{eq:sc1s} is
a relaxation of the standard (dense) block-sparse recovery one~\cite[Theorem~2]{EKB09}.
\end{theorem}

\begin{proofi}
  Recall that
  $\oinv\dictmGoc=\inv{\dictmSo\transp\pdictm\sblock{S_0}}\dictm\sblock{S_0}\transp\pdictm\sblock{\cpl{G}_0}$.
  Since $\rhoc{\cdot,\cdot}(\cdot)$ is submultiplicative,
  \cite{EKB09},\footnote{There is a slight abuse of notation here, in that,
    in our case of non-square blocks, each norm $\rhoc{\cdot,\cdot}(\cdot)$
    in the right hand of the submultiplicativity inequality~\refeq{eq:rhoc-submul}
    is actually a different norm. However, it is easy to see that the
    referred inequality holds in this case as well.}
\begin{eqnarray}
\label{eq:rhoc-submul}
\rhocSoGoc(\oinv\dictmGoc)& \leq &
\rhocSoSo((\dictmSo\transp\pdictm\sblock{S_0})^{-1})\,\rhocSoGoc(\dictmSo\transp\pdictm\sblock{\cpl{G}_0}).
\end{eqnarray}
Applying the definitions of $\rhocSoGoc$ and $\mup^s$ we have,
\begin{equation}
\label{eq:coso2}
\rhocSoGoc(\dictmSo\transp\pdictm\sblock{\cpl{G}_0}) 
=
\max_{F \in \cpl{\mathcal{G}}_0} \sum_{E \in \mathcal{S}_0} \rho(\dictm\sblock{E}\transp\pdictm\sblock{F})
\leq k \max_{F \in \cpl{\mathcal{G}}_0} \max_{E \in \mathcal{S}_0}\{\rho(\dictm\sblock{E}\transp\pdictm\sblock{F})\} 
\leq 
k g \mup^s,
\end{equation}
where the last inequality in \refeq{eq:coso2} derives from \refeq{eq:pco-2}
and the fact that each $E \in \mathcal{S}_0$ belongs to some $G_i$, and
$|E|=s$, thus playing the role of the set $T$ in the definition of
$\rho^s(\cdot)$.
Our goal is now to obtain a bound for
$\rhocSoSo((\dictmSo\transp\pdictm\sblock{S_0})^{-1})$. To this end, we
define $\mat{Z}=\dictmSo\transp\pdictm\sblock{S_0}$, and rewrite it as
$\mat{Z}=\scalm^{-1}(\mat{I}-(\mat{I}-\scalm\mat{Z}
\scalm))\scalm^{-1}$. Here $\scalm$ is a $ks{\times}ks$ block-diagonal
scaling matrix to be defined later. Assume for now that
$\rhocSoSo(\mat{I}-\scalm\mat{Z}\scalm) < 1$. This allows us to apply the
following result from~\cite{EKB09}:
\begin{lemma}
  \label{lemma:neumann} Suppose that
  $\rhoc{\mathcal{E},\mathcal{F}}(\bbw)<1$. Then
  $(\bbi+\bbw)^{-1}=\sum_{k=0}^\infty (-\bbw)^k$.
\end{lemma}
By applying Lemma~\ref{lemma:neumann} to
$\mat{W}=-\mat{I}+\scalm\mat{Z}\scalm$ we can write
$\mat{Z}^{-1} = \scalm \left[ \sum_{i=0}^\infty (\mat{I}-\scalm\mat{Z}\scalm)^i\right] \scalm.$ With this,
\begin{align}
\rhocSoSo(\mat{Z}^{-1}) 
& \stackrel{(a)}{\leq}
\left[\rhocSoSo(\scalm)\right]^2
\rhocSoSo\left(\sum_{i=0}^\infty (\mat{I}-\scalm\mat{Z}\scalm)^i\right)
\stackrel{(b)}{\leq}
\left[\rhocSoSo(\scalm)\right]^2
\sum_{i=0}^\infty \rhocSoSo\left((\mat{I}-\scalm\mat{Z}\scalm)^i\right)\nonumber\\
&\stackrel{(c)}{\leq}
\left[\rhocSoSo(\scalm)\right]^2
\sum_{i=0}^\infty \left(\rhocSoSo(\mat{I}-\scalm\mat{Z}\scalm)\right)^i
 \stackrel{(d)}{\leq}
\frac{[\rhocSoSo(\scalm)]^2}{{1-\rhocSoSo(\mat{I}-\scalm\mat{Z}\scalm)}},
\label{eq:rhoc-invZ}
\end{align}
where in $(a)$ and $(c)$ we applied the submultiplicativity of
$\rhoc{\cdot,\cdot}(\cdot)$, $(b)$ is a consequence if the triangle
inequality, and $(d)$ is the limit of the geometric series, which
is finite when  $\rhocSoSo(\mat{I}-\scalm\mat{Z}\scalm) < 1$.

We now bound $\rhocSoSo(\mat{I}-\scalm\mat{Z}\scalm)$.  First, note that,
since $\Lambda$ is block-diagonal, we have that
$(\mat{I}-\scalm\mat{Z}\scalm)\sblock{S_i,S_j}=\mat{I}\sblock{S_i,S_j}-\scalm\sblock{S_i,S_i}\mat{Z}\sblock{S_i,S_j}\scalm\sblock{S_j,S_j}$.
We then choose $\Lambda$ to be a diagonal matrix with
$\Lambda_{ii}=(\dictv_i\transp\pdictv_i)^{-1/2}, i \in S_0$. With this
choice, we have that the diagonal elements of
$\mat{I}\sblock{S_j,S_j}-\scalm\sblock{S_j,S_j}\mat{Z}\sblock{S_j,S_j}\scalm\sblock{S_j,S_j}$ are
equal to $1$ for all $j$, and the off-diagonal elements are bounded by
$\nup$. Using Ger\v{s}gorin Theorem we then have that
\begin{equation}
\label{eq:coso4}
\rho(\mat{I}\sblock{S_j,S_j}-\scalm\sblock{S_j,S_j}\mat{Z}\sblock{S_j,S_j}\scalm\sblock{S_j,S_j}) \leq (s-1)\nup,\,j=1,\ldots, k.
\end{equation}
As for the off-diagonal $s{\times}s$ blocks of $\mat{I}-\scalm\mat{Z}\scalm$, we have $(\mat{I}-\scalm\mat{Z}\scalm)\sblock{S_i,S_j}=-\scalm\sblock{S_i,S_i}\mat{Z}\sblock{S_i,S_j}\scalm\sblock{S_j,S_j}$. We then have
\begin{equation}
\label{eq:coso5}
\rho((\mat{I}-\scalm\mat{Z}\scalm)\sblock{S_i,S_j})
\stackrel{(a)}{\leq}
\rho(\scalm\sblock{S_i,S_i})\rho(\mat{Z}\sblock{S_i,S_j})\rho(\scalm\sblock{S_j,S_j}) 
\stackrel{(b)}{\leq} \zeta (g\mup^{ss}) \zeta,
\end{equation}
where in $(a)$ we used the submultiplicativity of $\rho(\cdot)$, and $(b)$
derives from the definition of $\mup^{ss}$, and the fact that, with our
choice of $\Lambda$ we have $\rho(\scalm\sblock{S_i,S_i}) \leq \zeta$ for
all $i$. Now we can write the definition of
$\rhocSoSo(\mat{I}-\scalm\mat{Z}\scalm)$ and bound its summation using
\refeq{eq:coso4}--\refeq{eq:coso5},
\begin{align}
\rhocSoSo(\mat{I}-\scalm\mat{Z}\scalm) 
&\leq 
\max_{S_j:j \leq k} \left\{
\rho(\mat{I}\sblock{S_j,S_j}-\scalm\sblock{S_j,S_j}\mat{Z}\sblock{S_j,S_j}\scalm\sblock{S_j,S_j} )\right. + \ldots \nonumber\\
& \ldots\left.\sum_{S_i:i \leq k, i \neq j}\!\!\!\!\rho\left(\scalm\sblock{S_i,S_i}(\mat{I}-\scalm\mat{Z}\scalm)\sblock{S_i,S_j}\scalm\sblock{S_j,S_j}\right) 
\right\} \leq (s-1)\nup+g\mup^{ss}\zeta^2.\label{eq:rho-I-DZD-bound}
\end{align}
By our choice of $\scalm$, $\rho(\scalm\sblock{S_i,S_i}) \leq \zeta$ and
$\rho(\scalm\sblock{S_i,S_j})=0$ for $i \neq j$. Therefore
$\rhocSoSo(\scalm) \leq \zeta$ as well. Using this together with
\refeq{eq:rho-I-DZD-bound} in \refeq{eq:rhoc-invZ}, we obtain
\begin{equation}
\label{eq:rhoc-invZ-bound}
\rhocSoSo(\mat{Z}^{-1}) \leq \frac{\zeta^2}{1-(s-1)\nup+g\mup^{ss}  (k-1)\zeta^2}.
\end{equation}
To ensure that $\rhocSoSo(\mat{I}-\scalm\mat{Z}\scalm) < 1$, we need the
denominator in the above equation to be positive. Now \refeq{eq:sc1s}
follows by plugging \refeq{eq:coso2} and \refeq{eq:rhoc-invZ-bound} into
\refeq{eq:rhoc-submul},
\[
\rhocSoGoc(\oinv\dictmGoc) \leq  \frac{\zeta^2 kg\mup^s}{1-(s-1)\nup+g\mup^{ss}  (k-1)\zeta^2}.
\]

Finally, we use the same ideas to bound $\|\pinv\dictmGoc\|_{1,1}$ and
derive (\ref{eq:sc3s}). Specifically,
\begin{equation}
\label{eq:norm11-submul}
\|\pinv\dictmGoc\|_{1,1}\leq
\|(\dictmSo\transp\dictmSo)^{-1}\|_{1,1}\|\dictmSo\transp\dictmGoc\|_{1,1}.
\end{equation}
Now
\begin{equation}
\label{eq:sc3-ineq1}
\|(\dictmSo)\transp\dictmGoc\|_{1,1}=\max_{j \in \cpl{G_0}}
\sum_{i \in S_0} |\dictv\transp_i \dictv_j| \stackrel{(a)}{\leq} ks\crossco,
\end{equation}
where $(a)$ follows from the definition of $\crossco$ and the fact that
$|S_0|=ks$. It remains to develop a bound on
$\|(\dictmSo\transp\dictmSo)^{-1}\|_{1,1}$. To this end we express
$\dictmSo\transp\dictmSo = \mat{I} + \mat{W}$, and bound
\begin{equation}
\label{eq:W11-bound}
\|\mat{W}\|_{1,1} = \max_{r \leq k} 
\left\{\max_{i \in S_r} \left\{ \sum_{j \in S_r, j \neq i} 
|\dictv_i\transp\dictv_j| + \sum_{j \in S_0 \setminus S_r} |\dictv_i\transp\dictv_j|\right\} \right\}
\leq (s-1)\subco + s(k-1)\crossco.
\end{equation}
since for all $S_r, r \leq k$, and all $i \in S_r$, the first sum has
$(s-1)$ nonzero elements bounded by $\nu$, and the second sum has $s(k-1)$
elements bounded by $\crossco$. Now, by requiring $(s-1)\subco +
s(k-1)\crossco<1$ we can apply Lemma~\ref{lemma:neumann} to $\mat{W}$ and
follow the same path as the one that leads to \refeq{eq:rho-I-DZD-bound}, now using
the matrix norm properties of $\norm{\cdot}_{1,1}$, to obtain,
\begin{equation}\label{eq:coso8}
\|(\dictmSo\transp\dictmSo)^{-1}\|_{1,1} \leq \frac{1}{1-(s-1)\subco + s(k-1)\crossco}.
\end{equation}
Again, $(s-1)\subco + s(k-1)\crossco<1$ is implicit in the requirement that
the above denominator be positive. Plugging \refeq{eq:coso8} and
\refeq{eq:sc3-ineq1} into \refeq{eq:norm11-submul} yields (\ref{eq:sc3s}).

The proof for \refeq{eq:sc4s} is analogous to that of \refeq{eq:sc3s}, only
that now the upper bound on $|\dictv_i\transp\dictv_j|,i \in S_0, j \in
T_0$, is $\subco \leq \mutualco$. Continuing as before leads to
\refeq{eq:sc4s}.\end{proofi}

Theorems~\ref{thm:sc} and~\ref{thm:mu} are for the non-collaborative
case. For the collaborative case there exist results that show that both the
C-Lasso \cite{ER10} and C-GLasso \cite{boufounos10} will recover the true
shared active set with a probability of error that vanishes exponentially
with $\nsamples$. Since the in-group active sets are not necessarily equal
for all samples in $\datam$, C-HiLasso could only help in recovering the
group sparsity pattern. Since the C-GLasso is a special case of C-HiLasso
when $\lambda_1=0$, we can conjecture that when $\lambda_1>0$, the accuracy
of the C-HiLasso in recovering the correct groups will improve with
larger $\nsamples$. Furthermore, since our results for HiLasso improve on
those of the Group Lasso, it is to be expected that the accuracy of
C-HiLasso, for an appropriate $\lambda_1>0$, will be better than that of
C-GLasso.

As an intuitive explanation to why this may happen, the proofs in
\cite{ER10} and \cite{boufounos10} assume a continuous probability
distribution on the non-zero coefficients of the signals, and give recovery
results for the average case. On the other hand, the in-group sparsity
assumption of C-HiLasso implies that only $s$ out of $\groupsize$ samples
will be nonzero within each group. This implies that, for the same group
sparsity pattern, there will be much less (exactly a fraction
$s/\groupsize$) non-zero elements in the possible signals compared to the
ones that can occur under the hypothesis of C-GLasso. Since any assumed
distribution of the signals under the in-group sparsity hypothesis has to be
concentrated on this much smaller set of possible signals, they should be
easier to recover correctly from solutions to the C-HiLasso program,
compared to the dense group case of C-GLasso.



\section{Experimental results}
\label{sec:results}

In this section we show the strength of the proposed HiLasso and C-HiLasso
models.  We start by comparing our model with the standard Lasso and Group
Lasso using synthetic data.  We created $\ngroups$ dictionaries, $\mat{D}_r,
r=1,\ldots,\ngroups$, with $\groupsize=64$ atoms of dimension $\ndims=64$,
and i.i.d. Gaussian entries. The columns were normalized to have unit
$\ell_2$ norm. We then randomly chose $k=2$ groups to be active at each time
(on all the signals).  Sets of $\nsamples=200$ normalized testing signals
were generated, one per active group, as linear combinations of $s \ll 64$
elements of the active dictionaries, $\mat{x}_j^r = \mat{D}_r
\coefv_j^r$. The mixtures were created by summing these signals and
(eventually) adding Gaussian noise of standard deviation $\sigma$. The
generated testing signals have a hierarchical sparsity structure and while
they share groups, they do not necessarily share the sparsity pattern inside
the groups. We then built a single dictionary by concatenating the
sub-dictionaries, $\mat{D} = [\mat{D}_1,\ldots,\mat{D}_\ngroups]$, and used
it to solve the Lasso, Group Lasso, HiLasso and C-HiLasso problems.
Table~\ref{tab:multi-signal-mse} summarizes the Mean Squared Error
(\textsc{mse}) and Hamming distance of the recovered coefficient vectors
$\coefv_j, j=1,\ldots,\nsamples$. We observe that our model is able to
exploit the hierarchical structure of the data as well as the collaborative
structure.  Group Lasso selects in general the correct blocks but it does
not give a sparse solution within them. On the other hand, Lasso gives a
solution that has nonzero elements belonging to groups that were not active
in the original signal, leading to a wrong model/class selection. HiLasso
gives a sparse solution that picks atoms from the correct groups but still
presents some minor mistakes. For the collaborative case, in all the tested
configurations, no coefficients were selected outside the correct active
groups, and the recovered coefficients are consistently the best ones.

In all the examples, and for each method, the regularization parameters were
the ones for which the best results were obtained.  One can scale the
parameter $\lambda_2$ to account for different number of signals. This
situation is analogous to a change in the size of the dictionary, thus,
$\lambda_2$ should be proportional to the square root of the number of
signals to code.

\newcommand{\best}[1]{{\color{blue}\bf #1}}
\begin{table*}
\begin{center}
{\footnotesize\setlength\tabcolsep{4pt}
\begin{tabular}{|c|cc|}\hline
$\sigma=0.1$ & 417 / 22.0          & 1173 / 361.6 \\
             & 330 / 19.8          &\best{163} / \best{13.3} \\ \hline
$\sigma=0.2$ & 564 / 21.6          & 1182 / 378.3 \\
             & 399 / 22.7          &\best{249} / \best{17.1}  \\ \hline
$\sigma=0.4$ & 965 / 22.7          & 1378 / 340.3 \\
             & 656 / \best{19.5} & \best{595}  / 27.4  \\ \hline
\end{tabular}\hspace{1ex}\begin{tabular}{|c|cc|}\hline
$s=8$   & 388 / 22.0          & 1184 / 318.2 \\
        & 272 / 19.5          & \best{96} / \best{16.2}\\ \hline
$s=12$  & 1200 / 36.2          & 1166 / 350.4 \\
        & 704 / \best{26.5} & \best{413} / 29.1\\ \hline
$s=16$  & 1641 / 43.9          & 1093 / 338.6 \\
        & 1100 / \best{32.2} & \best{551} / 35.0\\ \hline
\end{tabular}\hspace{1ex}\begin{tabular}{|c|cc|}\hline
$\ngroups=4$ & 1080 / 27.8          & 1916 / 221.7 \\
             & 1009 / \best{29.8} & \best{742} / 30.2\\ \hline
$\ngroups=8$ & 1200 / 36.2          & 1166 / 350.4 \\
             & 704 / \best{26.5} &\best{413} / 29.1\\ \hline
$\ngroups=12$& 1030 / 41.8          & 840 / 447.7 \\
             & 662 / \best{26.4} & \best{4} / 29.8\\ \hline
\end{tabular}
}
\end{center}
\caption{%
  \label{tab:multi-signal-mse}%
  \footnotesize%
  Simulated signal results. In every table, each $2{\times}2$ 
  cell contains the MSE ($\times10^{4}$) and Hamming distance (MSE/Hamming)
  for Lasso (top,left), GLasso (top,right), HiLasso (bottom,left) and C-HiLasso (bottom,right).
  In the first case (left) we vary the noise $\sigma$ while keeping
  $\ngroups=8$ and $s=8$ fixed. 
  In the second and third cases we have $\sigma=0$. For the second 
  experiment (center) we fixed $\ngroups = 8$ while changing $s$.
  In the third case we fix $s=12$ and vary the number of groups $\ngroups$.
  Bold blue indicates the best results, always obtained for the proposed models.
  In all cases, the number of active groups is $k=2$.
}
\end{table*}

We then experimented with the USPS digits dataset, which has been shown to
be well represented in the sparse modeling framework \cite{CVPR}. Here the
signals are vectors containing the unwrapped gray intensities of $16\times
16$ images ($\ndims=256$). We obtained each of the $\nsamples=200$ samples
in the testing data set as the mixture of two randomly chosen digits, one
from each of the two drawn sets of digits.  In this case we only have ground
truth at the group level. We measure the recovery performance in terms of
the average MSE of the recovered signals, $\mathrm{AMSE}=\frac{1}{\nsamples
  \ngroups} \sum_{r=1}^{\ngroups} \sum_{j=1}^{\nsamples} \norm{\mat{x}_j^r -
  \hat{\mat{x}}_j^r}_2^2$, where $\mat{x}_j^r$ is the component
corresponding to source $r$ in the signal $j$, and $\hat{\mat{x}}_j^r$ is
the recovered one.
%
%
 \begin{table*}[t]
 \begin{center}
{\footnotesize 
 \begin{tabular}{|c|cccccccccc|}\hline
 experiment         & \multicolumn{2}{c}{Lasso}     &  \multicolumn{2}{c}{GLasso}    & \multicolumn{2}{c}{HiLasso}    & \multicolumn{2}{c}{C-GLasso}  & \multicolumn{2}{c|}{C-HiLasso} \\\hline  \hline
                    & AMSE & Hamm & AMSE & Hamm & AMSE & Hamm & AMSE & Hamm & AMSE & Hamm  \\ 
1 digit   & 0.06&0.43 & 0.07&0.78 & 0.02&0.19 & \best{0.01}&\best{0.02} & 0.02&0.06 \\
1 digit+n & 0.08&1.31 & 0.08&0.87 & 0.04&0.48 & 0.05&0.25 & \best{0.02}&\best{0.01} \\
2 digit   & 0.09&1.46 & 0.08&1.86 & 0.02&1.18 & \best{0.01}&\best{0.74}& 0.02&0.90 \\
2 digit+n & 0.11&2.21 & 0.08&1.99 & 0.04&1.46 & 0.09&1.60 & \best{0.03}&\best{0.70} \\\hline
 \end{tabular}
}
  \end{center}
\vspace{-3ex}\caption{%
   \label{fig:digits}%
   \footnotesize%
   Noisy digit mixtures results.
   Four different cases are shown: when each signal is a single 
   digit and when it is the mixture of two different (randomly selected) digits,
   with and without additive Gaussian noise with standard deviation $10\%$ of the peak value. 
   For the 2 digits case, results
   are the average of 8 runs (in each round a new pair of digits was
   randomly selected). In the single digit case, the result is the
   average of the ten possible situations. 
   Both AMSE and Hamming distance are shown, with bold blue indicating best.
   Without noise, both C-GLasso and C-HiLasso yield very good results. However, in the noisy
   case, C-HiLasso is clearly superior, showing the advantage of adding regularization inside the groups
   from a robustness perspective. 
   See also Figure~\ref{fig:mixture3digits}.%
 }
 \end{table*}
 \begin{figure}
\begin{center}\includegraphics[width=0.4\textwidth]{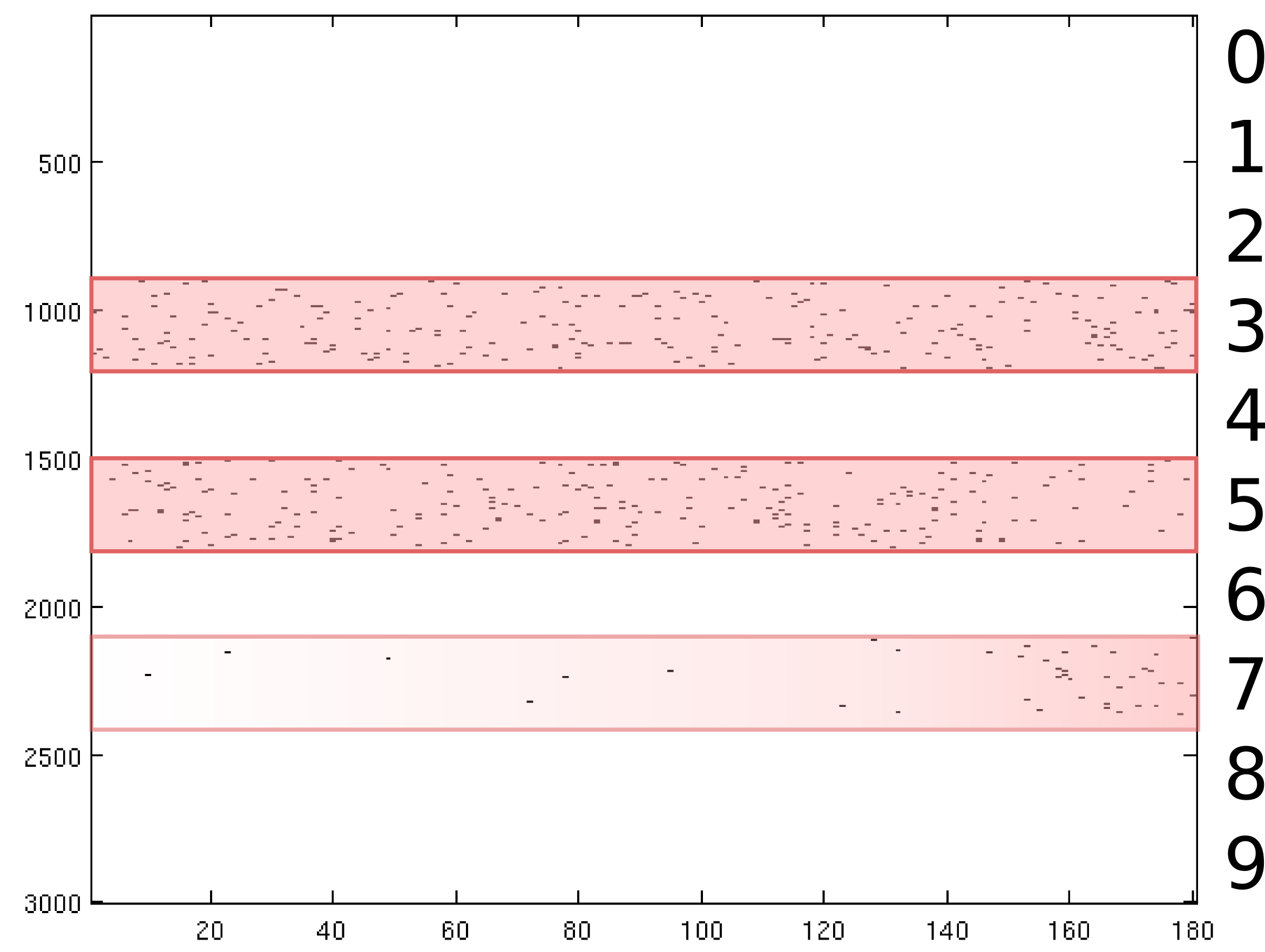}
\end{center}
\vspace{-1ex}\caption{\label{fig:mixture3digits} In this example we used
  C-HiLasso to analyze mixtures where the data set contains different number
  and types of sources/classes. We used a set containing 180 mixtures of
  digit images. The first 150 images are obtained as the sum/mixture of a
  number ``3'' and an number ``5'' (randomly selected). Each of the last 30
  images in the set are the mixture of three numbers: ``3'' ,``5'' and ``7''
  (the 180 images are of course presented at random, the algorithm is not
  a priori aware which images contain 2 sources and which contain 3). The
  figure shows the active sets of the recovered coefficients matrix $\coefm$
  as a binary matrix the same size as $\coefm$ (atom indices in the vertical
  and sample indices in the horizontal), where black dots indicate nonzero
  coefficients. C-HiLasso managed to identify the active blocks while the
  sub-dictionary corresponding to ``7'' is mostly active for the last 30
  images.  The accuracy of this result depends on the relationship between
  the sub-dictionaries corresponding to each digit.}
\end{figure}

Using the usual training-testing split for USPS, we first learned a
dictionary for each digit. We then created a single dictionary by
concatenating
them. In Table~\ref{fig:digits} we show the $\mathrm{AMSE}$ obtained while
summing $k=2$ different digits. We also consider the situation were only one
digit is present. C-HiLasso automatically detects the number of sources
while achieving the best recovery performance. As in the synthetic case,
only the collaborative method was able to successfully detect the true
active classes. In Figure~\ref{fig:mixture3digits} we relax the assumption
that all the signals have to contain exactly the same type and amount of
classes in the mixture, further demonstrating the flexibility of the
proposed C-HiLasso model.

\begin{figure*}[t]
\begin{center}
\includegraphics[height=0.12\textheight]{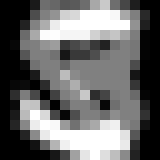} %
\includegraphics[height=0.12\textheight]{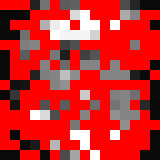} %
\includegraphics[height=0.12\textheight]{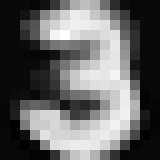} %
\includegraphics[width=0.12\textheight]{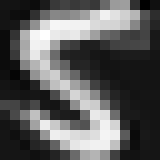} %
\includegraphics[height=0.12\textheight]{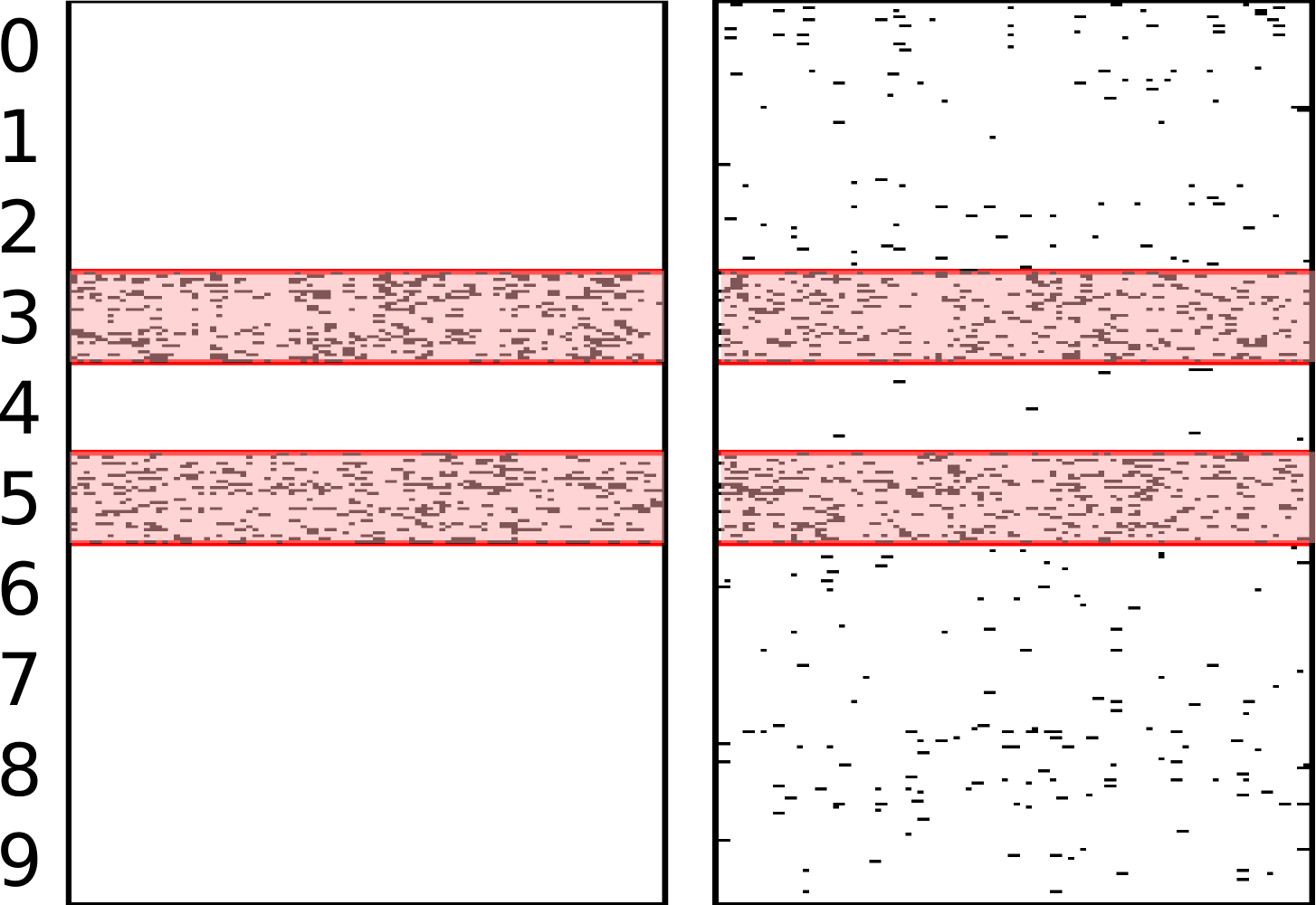} %
\end{center}
\vspace{-1ex}\caption{%
  \label{fig:missing}%
  \footnotesize%
  Example of recovered digits (3 and 5) from a mixture with 60\% of missing
  components. From left to right: noiseless mixture, observed mixture with
  missing pixels highlighted in red, recovered digits 3 and 5, and active
  set recovered for all samples using the C-HiLasso and Lasso
  respectively. In the last two figures, the active sets are represented as
  in Figure~\ref{fig:mixture3digits}. The coefficients blocks for digits 3
  and 5 are marked as pink bands.  Notice that the C-HiLasso exploits
  efficiently the hypothesis of collaborative group-sparsity, succeeding in
  recovering the correct active groups in all the samples. The Lasso, which
  lacks this prior knowledge, is clearly not capable of doing so, and active
  sets are spread over all the groups.}
\end{figure*}

We also used the digits dataset to experiment with missing data. We randomly
discarded an average of 60\% of the pixels per mixed image and then applied
C-Hilasso. The algorithm is capable of correctly detecting which digits are
present in the images. Some example results for this case are shown in
Figure~\ref{fig:missing}.  Note that this is a quite different problem than
the one commonly addressed in the matrix completion literature.  Here we do
not aim to recover signals that all belong to a unique unknown subspace, but
signals that are the combination of two non-unique spaces to be
automatically identified from the available dictionary. Such unknown spaces
have common models/groups for all the signals in question (the coarse level
of the hierarchy), but not necessarily the exact same atoms inside the
groups and therefore do not necessarily belong to the same subspaces. Both
levels of the hierarchy are automatically detected, e.g., the groups
corresponding to ``3'' and ``5,'' and the corresponding reconstructing atoms
(subspaces) in each group, these last ones possibly different for each
signal in the set.  While we consider that the possible subspaces are to be
selected from the provided dictionary (learned off-line from training data),
in Section~\ref{sec:discussion} we discuss learning such dictionaries as
part of the optimization as well (see also \cite{IMAapr2010, rosenblum10}). In such
cases, the standard matrix completion problem becomes a particular case of
the C-HiLasso framework (with a single group and all the signals having the
same active set, subspace, in the group), naturally opening numerous
theoretical questions for this new more general model.\footnote{Prof. Carin
  and collaborators have new results on the case of a single group and
  signals in possible different subspaces of the group, an intermediate
  model between standard matrix completion and C-HiLasso (personal
  communication).}

\begin{figure*}[t]
\begin{center}
\includegraphics[width=0.7\textwidth]{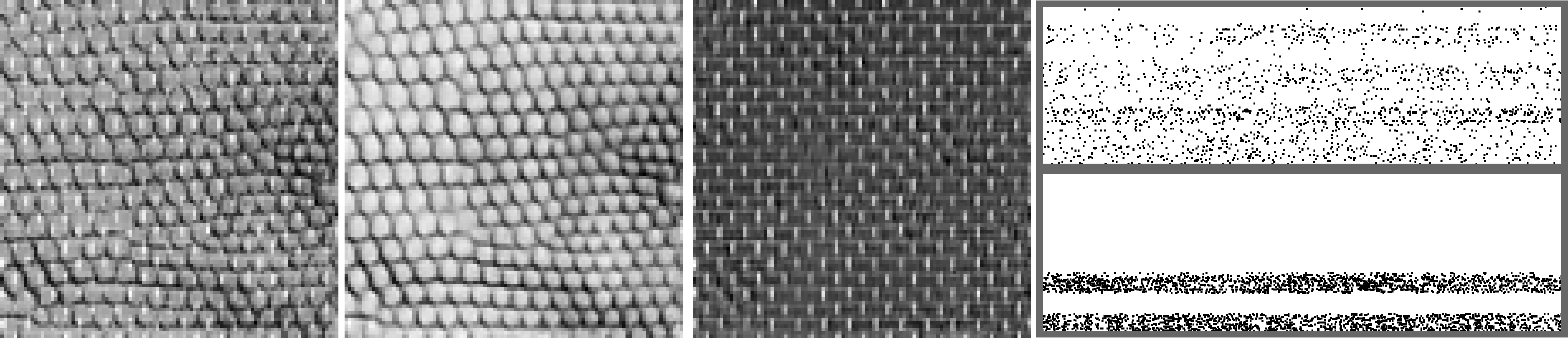}
\end{center}
\vspace{-1ex}\caption{\label{fig:texture}%
  \footnotesize%
  Texture separation results. Left to right: sample mixture,
  corresponding C-HiLasso separated textures, and  comparison
  of the active set diagrams obtained by the Lasso (as in Figure
  \ref{fig:missing}). The one for Lasso is shown on top, where all groups
  are wrongly active, and the one for C-HiLasso on bottom, showing that
  only the two correct groups are selected.}
\end{figure*}

We also compared the performance of C-HiLasso, Lasso, GLasso
and C-GLasso (without hierarchy) in the task of separating mixed textures in
an image. In this case, the set of signals $\datam$ corresponds to all
$12{\times}12$ patches in the (single) image to be analyzed.  We chose 8
textures from the Brodatz dataset and trained one dictionary for each one of
them using one half of the respective images (these form the $g=8$ groups of
the dictionary). Then we created an image as the sum of the other halves of
the $k=2$ textures. One can think of this experiment as a generalization to
the texture separation problem proposed in \cite{EladTexture} (without
additive noise), where only two textures are present. The experiment was
repeated for all possible combinations of two textures from the 8 possible
ones. The results are summarized in Table~\ref{tab:texture-summary}. A
detailed example is shown in Figure~\ref{fig:texture}. For each algorithm,
the best parameters were chosen using grid search, ensuring that those were
not in the edges of the grid. For Lasso and C-HiLasso the best $\lambda_1$
is $0.0625$. For GLasso and C-GLasso, the best $\lambda_2$ was,
respectively, $0.05$ and $75$ (for the collaborative setting, we
heuristically scale $\lambda_2$  with the number of signals
as $\sqrt{\nsamples}$. In this experiment, $\nsamples \approx 512^2$,
  leading to such large value of $\lambda_2$).  From
  Table~\ref{tab:texture-summary} we can conclude that the C-HiLasso is
  significantly better than the competing algorithms, both in the MSE of the
  recovered signals (we show the AMSE of recovering both active
  signals), and in the average Hamming distance between the recovered
  group-wise active sets and the true ones.  In the latter case we observe
  that, in many cases, the C-HiLasso active set recovery performance is
  perfect (Hamming distance $0$) or near perfect, whereas the other methods
  seldom approach a Hamming distance lower than $1$.

\begin{table*}
\begin{center}
\footnotesize
\setlength{\tabcolsep}{2pt}
\resizebox{0.7\textwidth}{!}{%
\begin{tabular}{|c|cc|cc|cc|cc|cc|cc|cc|cc|}\hline
   &
\multicolumn{2}{|c|}{\includegraphics[width=10ex]{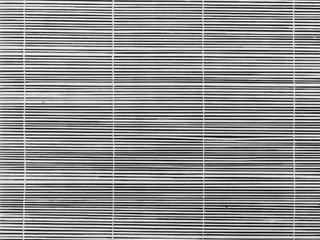}}&
\multicolumn{2}{|c|}{\includegraphics[width=10ex]{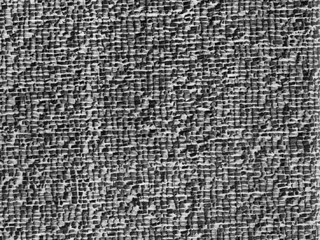}}&
\multicolumn{2}{|c|}{\includegraphics[width=10ex]{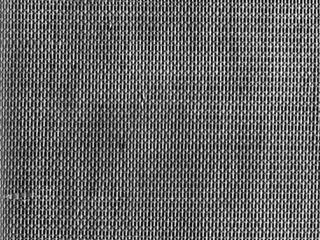}}&
\multicolumn{2}{|c|}{\includegraphics[width=10ex]{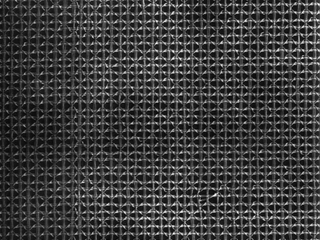}}&
\multicolumn{2}{|c|}{\includegraphics[width=10ex]{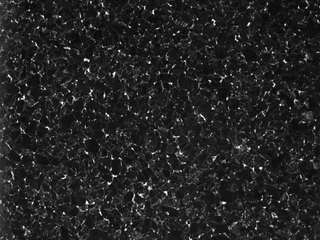}}&
\multicolumn{2}{|c|}{\includegraphics[width=10ex]{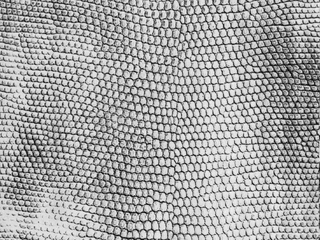}}&
\multicolumn{2}{|c|}{\includegraphics[width=10ex]{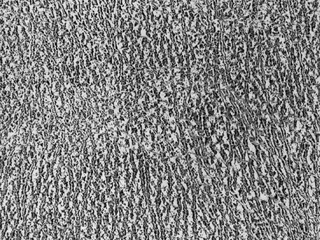}}&
\multicolumn{2}{|c|}{\includegraphics[width=10ex]{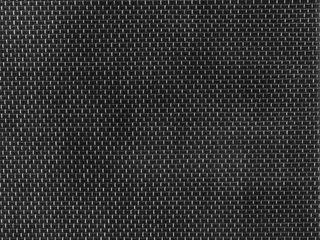}} \\ \hline
\multirow{2}{*}{\includegraphics[width=10ex]{tex49.png}} 
   &&&110&214&\best{18}&074& 63& 78& 19& 47& 85&174 &107&447&  7& 43 \\
   &&&117&\best{69}&069&\best{18}&126&\best{38}& 47&\best{18}&132&\best{51}&102&\best{42}& 27&\best{3} \\\hline
\multirow{2}{*}{\includegraphics[width=10ex]{tex84.png}} 
   &2.80&0.42&&& 107&  76& 141& 129&  91&  83& 191& 234& 240& 219&  68& 105 \\
   &1.36&\best{0.00}&&& 182&\best{ 68}&209&\best{102}&100&\best{ 78}&257&\best{141}&245&\best{178}& 95&\best{ 19} \\\hline
\multirow{2}{*}{\includegraphics[width=10ex]{tex53.png}} 
   &0.33&0.25&3.65&\best{0.00}&&&  52&\best{ 42}& 35& 62&105&112&162&141& 21& 93\\
   &2.06&\best{0.00}&2.67&0.02&&& 158& 43& 83&\best{ 29}&214&\best{ 62}&200&\best{107}&102&\best{ 10}\\\hline
\multirow{2}{*}{\includegraphics[width=10ex]{tex52.png}}
  &0.96&0.01&3.69&0.07&1.74&\best{0.00}&&&\best{ 49}& 72&123&145&182&148& 26& 89\\
  &1.97&\best{0.00}&2.30&\best{0.00}&2.42&\best{0.00}&&& 81& 55&224&\best{ 98}&214&\best{107}& 85&\best{ 10}\\\hline
\multirow{2}{*}{\includegraphics[width=10ex]{tex33.png}} 
  &1.02&1.00&3.55&1.00&1.42&1.00&2.25&1.00&&& 85& 76&120& 87& 15& 63\\
  &2.25&\best{0.09}&2.52&\best{0.94}&3.39&\best{0.16}&2.85&\best{0.35}&&&120& \best{59}&107& \best{71}& 41&  \best{9}\\\hline
\multirow{2}{*}{\includegraphics[width=10ex]{tex3.png}}
  &2.26&0.32&4.12&\best{0.53}&3.48&0.44&3.49&0.32&3.16&1.00&&&229&240& 56& 95\\
  &2.50&\best{0.00}&3.23&0.82&3.54&\best{0.20}&3.11&\best{0.01}&4.07&\best{0.40}&&&245&\best{162}&117& \best{27}\\\hline
\multirow{2}{*}{\includegraphics[width=10ex]{tex24.png}} 
  &4.37&1.39&4.47&\best{0.08}&4.09&0.13&4.23&0.12&4.20&1.00&4.42&0.42&&&100&112\\
  &2.51&\best{0.02}&2.39&0.22&2.42&\best{0.02}&2.76&\best{0.02}&2.24&\best{0.20}&2.96&\best{0.11}&&&102&\best{51} \\\hline
\multirow{2}{*}{\includegraphics[width=10ex]{tex6.png}}
  &0.09&0.98&3.77&1.00&0.31&1.00&1.83&1.00&1.13&1.00&3.14&0.97&4.30&1.00&& \\
  &0.53&\best{0.00}&1.75&\best{0.01}&2.04&\best{0.00}&1.82&\best{0.00}&2.18&\best{0.00}&3.04&\best{0.24}&1.90&\best{0.18}&& \\\hline
\end{tabular}
}
\end{center}
\caption{\label{tab:texture-summary}%
  Texture separation results. 
  The rows and columns indicate the active textures in each cell. 
  The upper triangle contains the AMSE ($\times10^{4}$) results, while the lower 
  triangle shows the Hamming error in the group-wise active set 
  recovery. Within each cell, results are shown for the Lasso 
  (top left), Group Lasso (bottom left), Collaborative Group Lasso 
  (top right) and Collaborative Hierarchical Lasso (bottom right). 
  The best results are in blue bold. 
  Note that, both for the AMSE and Hamming distance, in 26 out of 28 cases,
  our model outperforms previous ones.
}
\end{table*}
\label{obs:optimization2}Finally, we use C-HiLasso to automatically identify
the sources present in a mixture of audio signals \cite{audio}. The goal is
to identify the speakers talking simultaneously on a single recording.  Here
the task is not to fully reconstruct each of the unmixed sources from the
observed signal but to identify which speakers are active. In this case,
since the original sources do not need to be recovered, the modeling can be
done in terms of features extracted from the original signals in a linear
but non-bijective way.

Audio signals have in general very rich structures and their properties
rapidly change over time.  A natural approach is to decompose them into a
set of overlapping local time-windows, where the properties of the signal
remain stable. There is a straightforward analogy with the approach
explained above for the texture segmentation case, where images were
decomposed into collections of overlapping patches. These time-windows will
collaborate in the identification.

A challenging aspect when identifying audio sources is to obtain features
that are specific to each source and at the same time invariant to changes
in the fundamental frequency (pitch) of the sources.  In the case of speech,
a common choice is to use the short-term power spectrum envelopes as feature
vectors \cite{Speech} (refer to \cite{audio} for details on the feature
extraction process and implementation). The spectral envelope in human
speech varies along time, producing different patterns for each
phoneme. Thus, a speaker does not produce an unique spectral envelope, but a
set of spectral envelopes that live in a union of manifolds. Since such
manifolds are well represented by sparse models, the problem of speaker
identification is well suited for the proposed C-HiLasso framework, where
each block in the dictionary is trained for the features corresponding to a
given speaker, and the overlapping time-windows collaborate in detecting the
active blocks.
\begin{figure}
\begin{center}
\includegraphics[height=0.11\textheight]{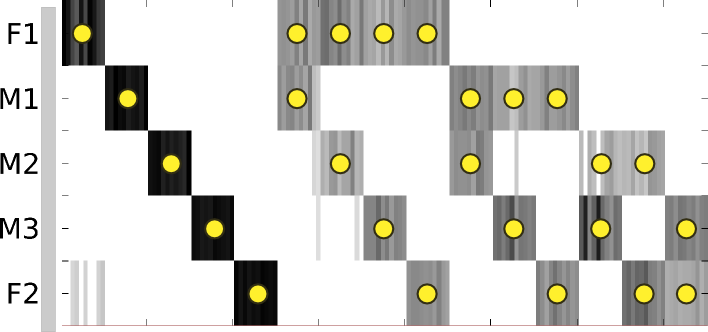}%
\caption{\label{fig:audio-results} Speaker identification results. Each
  column corresponds to the sources identified for a specific time frame,
  the true ones marked by yellow dots. The vertical axis indicates the
  estimated activity of the different sources, where darker colors indicate
  higher energy.  For each possible combination of speakers, 10 frames (15
  seconds of audio) were evaluated.}
\end{center}
\end{figure}

For this experiment we use a dataset consisting of recordings of five
different German radio speakers, two female and three male.  Each recording
is six minutes long. One quarter of the samples were used for dictionary
training, and the rest for testing.  For each speaker, we learned a
sub-dictionary from the training dataset.  For testing, we extracted $10$
non-overlapping frames of 15 seconds each (including silences made by the
speakers while talking), and encoded them using C-HiLasso. The experiment
was repeated for all possible combinations of two speakers, and all the
speakers talking alone.  The results are presented in
Figure~\ref{fig:audio-results}. C-HiLasso manages to detect automatically
the number of sources very accurately, as well as the actual active
speakers. Again, refer to \cite{audio} for comparisons with other sparse
modeling methods (showing the clear advantage of C-HiLasso) and results
obtained for the identification of wind instruments in musical
recordings.
%
%
%
\section{Discussion} 
\label{sec:discussion} We introduced a new framework of collaborative
hierarchical sparse coding, where multiple signals collaborate in their
encoding, sharing code groups (models) and having (possible disjoint) sparse
representations inside the corresponding groups. An efficient optimization
approach was developed, which guarantees convergence to the global minimum,
and examples illustrating the power of this framework were presented. At the
practical level, we are currently continuing our work on the applications of
this proposed framework in a number of directions, including collaborative
instruments separation in music, signal classification, and speaker
recognition, following the here demonstrated capability to collectively
select the correct groups/models.

At the theoretical level, a whole family of new problems is opened by this
proposed framework, some of which we already addressed in this work. A
critical one is the overall capability of selecting the correct groups in
the collaborative scenario, with missing information, and thereby of
performing correct model selection and source identification and separation.
Results in this direction will be reported in the future.

Finally, we have also developed an initial framework for learning the
dictionary for collaborative hierarchical sparse coding, meaning the
optimization is simultaneously on the dictionary and the code. As it is the
case with standard dictionary learning, this is expected to lead to
significant performance improvements (see \cite{CVPR} for the particular
case of this with a single group active at a time).

\small
\noindent{\bf Acknowledgments:} Work partially supported by NSF, NSSEFF,
ONR, NGA, and ARO. We thank Dr. Tristan Nguyen, when we presented him this
model, he motivated us to think in a hierarchical fashion and to look at
this as just the particular case of a fully hierarchical sparse coding
framework. 
We thank Prof. Tom Luo and Gonzalo Mateos for invaluable
  help on optimization methods.
  We thank Prof. Larry Carin, Dr. Guoshen Yu
and Alexey Castrodad for very stimulating conversations, and for the fact
that their own work (for LC, GY, and AC) also motivated in part the example
with missing information. The anonymous reviewers prompted an early mistake
in the proof of Theorem~1, and that, together with their additional
comments, led to improving the bounds in the theorem, as well as the overall
presentation of the paper. We also want to thank the reviewer for the closed
form inner loop of the proposed optimization method, which simplified it and
resulted in significant practical improvements.


\bibliography{hilasso}

\begin{thebibliography}{10}

\bibitem{yuan06}
M.~Yuan and Y.~Lin,
\newblock ``Model selection and estimation in regression with grouped
  variables,''
\newblock {\em J. Royal Stat. Society, Series B}, vol. 68, pp. 49--67, 2006.

\bibitem{bach09}
R.~Jenatton, J.~Audibert, and F.~Bach,
\newblock ``Structured variable selection with sparsity-inducing norms,''
  arXiv:0904.3523v1, 2009.

\bibitem{EM09}
Y.~C. Eldar and M.~Mishali,
\newblock ``Robust recovery of signals from a structured union of subspaces,''
\newblock {\em IEEE Trans. Inform. Theory}, vol. 55, no. 11, pp. 5302--5316,
  Nov. 2009.

\bibitem{tropp06a}
J.~Tropp,
\newblock ``Algorithms for simultaneous sparse approximation. part {II}: Convex
  relaxation,''
\newblock {\em Signal Processing}, vol. 86, no. 3, pp. 589--602, 2006.

\bibitem{tropp06b}
J.~Tropp, A.~Gilbert, and M.~Strauss,
\newblock ``Algorithms for simultaneous sparse approximation. part {I}: Greedy
  pursuit,''
\newblock {\em Signal Processing}, vol. 86, no. 3, pp. 572--588, 2006.

\bibitem{CREK05}
S.~Cotter, B.~Rao, K.~Engan, and K.~Kreutz-Delgado,
\newblock ``Sparse solutions to linear inverse problems with multiple
  measurement vectors,''
\newblock {\em IEEE Trans. Sig. Proc.}, vol. 53, no. 7, pp. 2477--2488, July
  2005.

\bibitem{CH06}
J.~Chen and X.~Huo,
\newblock ``Theoretical results on sparse representations of
  multiple-measurement vectors,''
\newblock {\em IEEE Trans. Sig. Proc.}, vol. 54, no. 12, pp. 4634--4643, Dec.
  2006.

\bibitem{ME08a}
M.~Mishali and Y.~C. Eldar,
\newblock ``Reduce and boost: {R}ecovering arbitrary sets of jointly sparse
  vectors,''
\newblock {\em IEEE Trans. Sig. Proc.}, vol. 56, no. 10, pp. 4692--4702, Oct.
  2008.

\bibitem{ER10}
Y.~C. Eldar and H.~Rauhut,
\newblock ``Average case analysis of multichannel sparse recovery using convex
  relaxation,''
\newblock {\em {I}{E}{E}{E} {T}rans. {I}nform. {T}heory}, vol. 56, no. 1, pp.
  505--519, 2010.

\bibitem{nesterov07}
Y.~Nesterov,
\newblock ``Gradient methods for minimizing composite objective function,''
\newblock in {\em {CORE} Discussion Paper 2007/76, Center for Operations
  Research and Econometrics ({CORE})}. Catholic University of Louvain,
  Louvain-la-Neuve, Belgium, 2007.

\bibitem{SpaRSA}
S.~Wright, R.~Nowak, and M.~Figueiredo,
\newblock ``Sparse reconstruction by separable approximation,''
\newblock {\em IEEE Trans. Sig. Proc.}, vol. 57, no. 7, pp. 2479--2493, 2009.

\bibitem{friedman10a}
J.~Friedman, T.~Hastie, and R.~Tibshirani,
\newblock ``A note on the group lasso and a sparse group lasso,''
\newblock preprint (2010), available at {\footnotesize
  \texttt{http://www-stat.stanford.edu/\texttildelow tibs}}.

\bibitem{peng}
J.~Peng, J.~Zhu, A.~Bergamaschi, W.~Han, D.~Noh, J.~Pollack, and P.~Wang,
\newblock ``Regularized multivariate regression for identifying master
  predictors with application to integrative genomics study of breast cancer,''
\newblock {\em Annals of Applied Statistics}, vol. 4, no. 1, pp. 53--77, 2010.

\bibitem{KimICML}
S.~Kim and E.~P. Xing,
\newblock ``Tree-guided group lasso for multi-task regression with structured
  sparsity,''
\newblock in {\em ICML}, June 2010.

\bibitem{Jenatton2010}
R.~Jenatton, J.~Mairal, G.~Obozinski, and F.~Bach,
\newblock ``Proximal methods for sparse hierarchical dictionary learning,''
\newblock in {\em ICML}, June 2010.

\bibitem{Starck04imagedecomposition}
J.~Starck, M.~Elad, and D.~Donoho,
\newblock ``Image decomposition via the combination of sparse representations
  and a variational approach,''
\newblock {\em IEEE Trans. Image Proc.}, vol. 14, pp. 1570--1582, 2004.

\bibitem{JenattonBach}
R.~Jenatton, J.~Mairal, G.~Obozinski, and F.~Bach,
\newblock ``Proximal methods for hierarchical sparse coding,''
\newblock Tech. {R}ep. HAL : inria-00516723, {INRIA}, 2010.

\bibitem{tibshirani96}
R.~Tibshirani,
\newblock ``Regression shrinkage and selection via the {LASSO},''
\newblock {\em J. Royal Stat. Society: Series B}, vol. 58, no. 1, pp. 267--288,
  1996.

\bibitem{CDS99}
S.~Chen, D.~Donoho, and M.~Saunders,
\newblock ``Atomic decomposition by basis pursuit,''
\newblock {\em SIAM J. Scientific Computing}, vol. 20, no. 1, pp. 33--–61,
  1999.

\bibitem{D06}
D.~Donoho,
\newblock ``Compressed sensing,''
\newblock {\em {IEEE} Trans. on Inf. Theory}, vol. 52, no. 4, pp. 1289--1306,
  Apr 2006.

\bibitem{giryes10}
R.~Giryes, M.~Elad, and Y.~C. Eldar,
\newblock ``The projected {GSURE} for automatic parameter tuning in iterative
  shrinkage methods,''
\newblock Submitted to {\em Applied and Computational Harmonic Analysis}, 2010.

\bibitem{nachoMDL}
I.~Ram\'{i}rez and G.~Sapiro,
\newblock ``Sparse coding and dictionary learning based on the {MDL}
  principle,'' arXiv:1010.4751, 2010.

\bibitem{zhouNIPS}
M.~Zhou, H.~Chen, J.~Paisley, L.~Ren, G.~Sapiro, and L.~Carin,
\newblock ``Non-parametric bayesian dictionary learning for sparse image
  representations,''
\newblock in {\em Proceedings of Advances in Neural Information Processing
  Systems (NIPS)}, 2009.

\bibitem{wright04}
B.~Turlach, W.~Venables, and S.~Wright,
\newblock ``Simultaneous variable selection,''
\newblock {\em Technometrics}, vol. 27, pp. 349--363, 2004.

\bibitem{CISS2010}
P.~Sprechmann, I.~Ramirez, and G.~Sapiro,
\newblock ``Collaborative hierarchical sparse modeling,''
\newblock in {\em CISS}, Mar. 2010.

\bibitem{boufounos10}
P.~Boufounos, G.~Kutyniok, and H.~Rauhut,
\newblock ``Sparse recovery from combined fusion frame measurements,''
  arXiv:0912.4988v1, 2010.

\bibitem{EKB09}
Y.~C. Eldar, P.~Kuppinger, and H.~B\"olcskei,
\newblock ``Block-sparse signals: {U}ncertainty relations and efficient
  recovery,''
\newblock {\em IEEE Trans. SP}, vol. 58, no. 6, pp. 3042--3054, June 2010.

\bibitem{daubechies04}
I.~Daubechies, M.~Defrise, and C.~De Mol,
\newblock ``An iterative thresholding algorithm for linear inverse problems
  with a sparsity constraint,''
\newblock {\em Comm. on Pure and Applied Mathematics}, vol. 57, pp. 1413--1457,
  2004.

\bibitem{CRT06}
E.~Cand\`es, J.~Romberg, and T.~Tao,
\newblock ``Robust uncertainty principles: {E}xact signal reconstruction from
  highly incomplete frequency information,''
\newblock {\em IEEE Trans. Inform. Theory}, vol. 52, no. 2, pp. 489--509, Feb.
  2006.

\bibitem{C08}
E.~Cand{\`e}s,
\newblock ``{T}he restricted isometry property and its implications for
  compressed sensing,''
\newblock {\em {C}. {R}. {A}cad. {S}ci. {P}aris {S}'er. {I} {M}ath.}, vol. 346,
  pp. 589--592, 2008.

\bibitem{tropp04}
J.~Tropp,
\newblock ``Greed is good: {A}lgorithmic results for sparse approximation,''
\newblock {\em IEEE Trans. Inform. Theory}, vol. 50, no. 10, pp. 2231--2242,
  Oct. 2004.

\bibitem{stojnic09}
M.~Stojnic,
\newblock ``Block-length dependent thresholds in block-sparse compressed
  sensing,'' arXiv:0907.3679, July 2009.

\bibitem{AEJL04}
A.~d'Aspremont, L.~El Ghaoui, M.~Jordan, and G.~Lanckriet,
\newblock ``A direct formulation for sparse {PCA} using semidefinite
  programming,''
\newblock {\em Neural Information Processing Systems}, vol. 17, 2004.

\bibitem{ZHT03}
H.~Zou, T.~Hastie, and R.~Tibshirani,
\newblock ``Sparse principal component analysis,''
\newblock {\em Journal of Computational and Graphical Statistics}, vol. 15, no.
  2, 2003.

\bibitem{MWA06}
B.~Moghaddam, Y.~Weiss, and S.~Avidan,
\newblock ``Spectral bounds for sparse {PCA}: Exact \& greedy algorithms,''
\newblock {\em Neural Information Processing Systems}, vol. 18, 2006.

\bibitem{CVPR}
I.~Ramirez, P.~Sprechmann, and G.~Sapiro,
\newblock ``Classification and clustering via dictionary learning with
  structured incoherence,''
\newblock in {\em CVPR}, June 2010.

\bibitem{IMAapr2010}
M.~Zhou, H.~Chen, J.~Paisley, L.~Ren, L.~Li, Z.~Xing, D.~Dunson, G.~Sapiro, and
  Lawrence Carin,
\newblock ``Nonparametric bayesian dictionary learning for analysis of noisy
  and incomplete images,''
\newblock IMA Preprint, April 2010, {\footnotesize
  \texttt{http://www.ima.umn.edu/preprints/apr2010/2307.pdf}}.

\bibitem{rosenblum10}
K.~Rosenblum, L.~Zelnik-Manor, and Y.~C. Eldar,
\newblock ``Sensing matrix optimization for block-sparse decoding,''
  arXiv:1009.1533, Sep. 2010.

\bibitem{EladTexture}
N.~Shoham and M.~Elad,
\newblock ``Alternating {KSVD}-denoising for texture separation,''
\newblock in {\em The IEEE 25-th Convention of Electrical and Electronics
  Engineers in Israel}, 2008.

\bibitem{audio}
P.~Sprechmann, I.~Ramirez, P.~Cancela, and G.~Sapiro,
\newblock ``Collaborative sources identification in mixed signals via
  hierarchical sparse modeling,'' arXiv:1010.4893, 2010.

\bibitem{Speech}
L.~Rabiner and B.-H. Juang,
\newblock {\em Fundamentals of {S}peech {R}ecognition},
\newblock Prentice-Hall, Inc., Upper Saddle River, NJ, USA, 1993.

\end{thebibliography}
\bibliographystyle{IEEEbib}

\end{document}